\documentclass{isprs}
\usepackage[T1]{fontenc}
\usepackage[utf8]{inputenc}
\usepackage{subfigure}
\usepackage{graphicx}
\usepackage{setspace}
\usepackage{geometry}
\usepackage{epstopdf}
\usepackage[labelsep=period]{caption}
\usepackage[british]{babel}
\usepackage[hang]{footmisc}
\usepackage{placeins}
\usepackage[table]{xcolor}
\usepackage{amsmath,amssymb,graphicx,booktabs,multirow,array,url,hyperref}
\usepackage{tabularx}

\newcommand{\TableSetup}{%
  \footnotesize  
  \setlength{\tabcolsep}{3.0pt}%
  \renewcommand{\arraystretch}{1.05}%
}

\graphicspath{{figures/}}
\begin{document}

\title{UAVFF3D: A Geometry-Aware Benchmark for Feed-Forward UAV 3D Reconstruction}

\date{}
\author{
Xiang Yang\textsuperscript{a,$\dagger$},
Yongli Wang\textsuperscript{a,$\dagger$},
HaiFeng Li\textsuperscript{a},
Yunsheng Zhang\textsuperscript{a,*}
}

\address{
\textsuperscript{a}School of Geosciences and Info-Physics, Central South University, Changsha, China\\
\textsuperscript{$\dagger$}These authors contributed equally to this work.\\
\textsuperscript{*}Corresponding author. Email: \texttt{zhangys@csu.edu.cn}
}

\commission{XX, }{YY}
\workinggroup{XX/YY}
\icwg{}

\abstract{
Feed-forward 3D reconstruction has advanced rapidly in recent years, yet existing models remain unreliable under UAV photogrammetric acquisition settings.
We argue that this unreliability cannot be attributed solely to appearance-domain shift, but is also driven by UAV-specific camera-geometry variations, particularly oblique viewing and HFOV--height ambiguity.
Existing UAV datasets often emphasize scene diversity while providing limited variation in UAV camera configurations, making them insufficient for robustness evaluation under camera-geometry changes and for UAV-domain adaptation.
To bridge this gap, we construct UAVFF3D, a geometry-aware real--synthetic benchmark for feed-forward UAV 3D reconstruction.
The dataset contains more than 170k real UAV images and more than 370k images synthesized from high-quality textured 3D models, covering a broad range of UAV camera-geometry configurations.
We also build a challenging test subset to diagnose model behavior under HFOV--height ambiguity.
Based on this dataset, we design a new evaluation protocol that jointly assesses camera-geometry estimation and reconstructed dense scene geometry under a shared global alignment, thereby reducing the bias introduced by separate alignments in existing evaluations.
Experiments on four representative feed-forward reconstruction models show that UAV-domain adaptation using UAVFF3D consistently improves both camera-geometry estimation and reconstructed dense scene geometry, reducing Ray Error by up to 84.2\%, Pose ATE by up to 76.0\%, and CD by up to 41.1\%.
In oblique-view scenes, domain adaptation substantially mitigates rotation-estimation degradation and reduces the oblique--nadir rotation gap by up to 90.7\%.
Under HFOV--height ambiguity, domain adaptation improves robustness across HFOV--height configurations and yields more stable performance across HFOV settings.
Incorporating camera priors further improves reconstruction performance under UAV-specific acquisition geometries.
The dataset and evaluation code are available on the project page: \url{https://github.com/yanxian-ll/UAVFF3D}.
}

\keywords{feed-forward reconstruction, UAV benchmark, HFOV--height ambiguity, oblique view}

\maketitle
\sloppy


\section{Introduction}
Feed-forward 3D reconstruction has recently emerged as an efficient alternative to traditional SfM/MVS pipelines
\cite{dust3r2024,mast3r2024,vggt2025,mapanything2025,depthanything3_2025,fast3r2025,cut3r2025,feedforward3d_survey2025}.
Instead of recovering 3D geometry through feature matching, bundle adjustment, dense matching, and fusion, these models directly predict camera parameters, depth, rays, point maps, or dense 3D structures from images.
This paradigm is particularly attractive for UAV photogrammetry, where applications such as urban modeling, disaster response, infrastructure inspection, low-altitude navigation, and digital-twin updating require rapid 3D reconstruction with accurate camera geometry and consistent scene structure
\cite{remondino2011uav_mapping,colomina2014uas_review,nex2014uav,jiang2021uav_3d_mapping_survey,hu2023uav_city_modeling,somanath2024urban_digital_twins,wang2025autonomous_uav_reconstruction}.

Despite this potential, current feed-forward reconstruction models remain unreliable when applied to UAV imagery.
A common interpretation is that this degradation arises from appearance-domain shift, a factor that has been widely discussed in UAV vision studies
~\cite{lyu2020uavid,gritzner2020using}.
However, failures in feed-forward 3D reconstruction cannot be explained by appearance shift alone.
Recent feed-forward reconstruction models are trained on large-scale and heterogeneous image collections, some of which already include UAV imagery or similar overhead data
~\cite{vggt2025,mapanything2025}.
Nevertheless, these models may still fail under specific UAV acquisition settings.
This observation suggests that the limitation is not merely a lack of UAV-like visual content in the training data. 
Rather, existing datasets provide insufficient and unsystematic coverage of the camera-geometry configurations required for UAV reconstruction.
We therefore argue that camera-geometry shift is a key source of failure in feed-forward UAV reconstruction.
Although general reconstruction datasets provide broad scene content, existing UAV datasets are often constrained by fixed sensor configurations, limited flight-altitude ranges, and restricted acquisition patterns. As a result, they cannot systematically cover the camera-geometry distribution required for feed-forward UAV reconstruction.
These factors directly affect camera rays, metric scale, depth distribution, multi-view correspondences, and camera--scene consistency.
Consequently, a model may perform well on visually similar UAV images but fail when the HFOV, flight altitude, viewing direction, or acquisition pattern changes.

\begin{figure}[t]
\centering
\includegraphics[width=\linewidth]{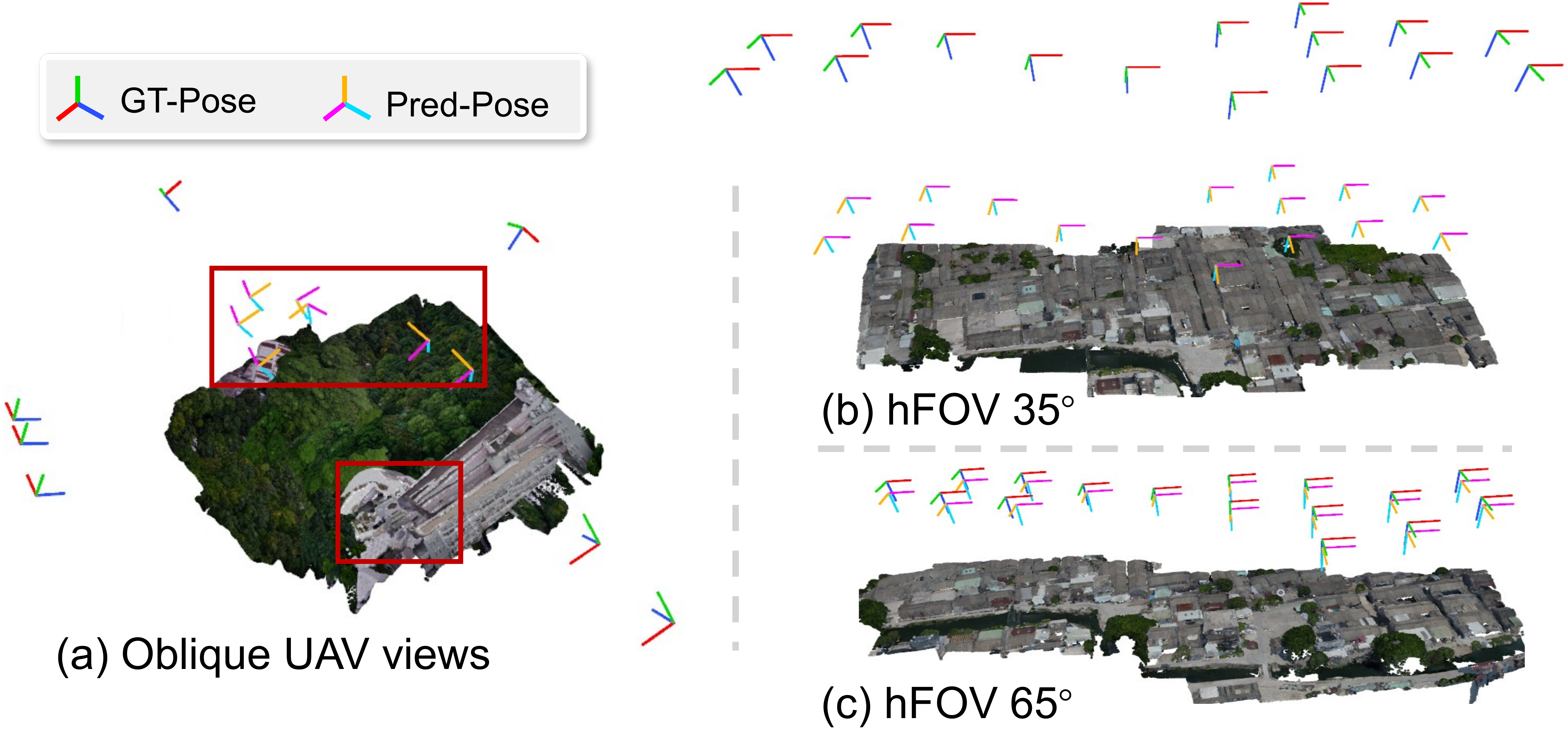}
\caption{
Typical failure cases of feed-forward UAV reconstruction.
(a) Oblique inputs produce incorrect poses and misaligned point clouds.
(b)--(c) Under similar image footprints, HFOV \(65^\circ\) yields better pose reconstruction than HFOV \(35^\circ\).
}
\label{fig:motivation_oblique_fov}
\end{figure}

This paper focuses on two representative camera-geometry challenges that are especially important for feed-forward UAV reconstruction: oblique-view degradation and HFOV--height ambiguity.
These two challenges correspond to viewpoint variation and projection-geometry variation in UAV acquisition, respectively, 
both of which can disrupt the consistent prediction of camera poses, scale, and dense scene structure by feed-forward models.
Figure~\ref{fig:motivation_oblique_fov} illustrates these two typical failure modes.

Oblique-view degradation is primarily associated with variations in UAV viewing direction.
Compared with near-nadir imagery, oblique UAV imagery exhibits greater perspective variation, stronger occlusion and disocclusion effects, more visible facades, repetitive building textures, and larger depth variation.
These factors make cross-view correspondence, rotation estimation, and dense geometry recovery more difficult.
As a result, a feed-forward model may predict locally plausible surfaces while placing cameras incorrectly or generating misaligned point clouds.
This indicates that UAV reconstruction failure is not merely a local depth-estimation problem, but a coupled problem of camera--scene consistency.

HFOV--height ambiguity arises from the coupling between field of view and flight altitude in UAV projection geometry.
In UAV imaging, the observed ground footprint is jointly determined by the camera field of view and the flight altitude.
A high-altitude narrow-FOV camera can produce image content similar to that of a low-altitude wide-FOV camera.
Although the images may appear visually similar, their underlying projection geometry differs: focal length, camera rays, camera height, and metric depth scale all change.
Therefore, RGB-only feed-forward models may rely on learned camera priors and struggle to infer the correct projection geometry and metric scale from image content alone.

Existing multi-view and UAV datasets provide important foundations for reconstruction, depth estimation, and urban modeling
\cite{dtu2016,eth3d2017,tankstemples2017,yao2020blendedmvs,liu2020whumvs,liu2023whuomvs,lin2021urbanscene3d,nex2024usegeo,nex2024usegeo_dataset,uavscenes2025,airzoo2026}.
However, these datasets are primarily designed to provide scene content, reconstruction supervision, or city-scale simulation, rather than to systematically evaluate the robustness of feed-forward models under changes in UAV camera geometry.
General multi-view datasets typically lack UAV-specific flight altitudes, HFOVs, and nadir/oblique acquisition patterns. 
Although existing UAV datasets are closer to the target domain, they are often restricted to fixed sensor configurations, narrow altitude ranges, and specific flight-route designs.
Thus, they are unable to cover the camera-geometry distribution required for feed-forward UAV reconstruction, especially diverse HFOV--height combinations and viewing-angle variations.

To address this gap, we propose UAVFF3D, a geometry-aware real--synthetic benchmark for feed-forward UAV 3D reconstruction.
The design goal of UAVFF3D is not merely to increase the number of scenes, but to systematically cover the camera-geometry variations that are critical in UAV reconstruction.
To this end, we construct synthetic UAV data with rich camera-geometry coverage, including diverse HFOVs, flight altitudes, viewing directions, and acquisition patterns, to support adaptation to UAV-specific projection and viewpoint distributions.
At the same time, we collect a large number of real UAV images so that domain adaptation does not rely solely on synthetic supervision and can also account for real-world appearance, noise, and acquisition-pattern differences.
In addition to training and adaptation data, UAVFF3D provides test data for geometric diagnosis and real-world evaluation.
To our knowledge, no existing UAV dataset provides controlled HFOV--height settings, in which projection geometry is varied while the image footprint remains approximately unchanged.
We therefore construct a controlled and challenging HFOV--height test set to evaluate the effect of projection-geometry changes on feed-forward reconstruction.
In addition, we build a LiDAR-supported real UAV test set that covers rich real acquisition geometries and enables evaluation of metric reconstruction capability under real-world conditions.
Based on these data, we further propose a shared scene-level alignment protocol that jointly evaluates camera predictions and reconstructed dense scene geometry under the same global coordinate transformation, thereby preventing separate alignments from masking camera--scene inconsistency.

The contributions of this paper are as follows:
\begin{itemize}
    \item \textbf{A geometry-aware real--synthetic UAV benchmark.}
    We propose UAVFF3D, a geometry-aware benchmark for feed-forward UAV 3D reconstruction.
    The benchmark combines large-scale real UAV imagery, synthetic data with rich camera-geometry coverage, a LiDAR-supported real test set, and a controlled HFOV--height test set. 
    It supports UAV-domain adaptation, real-world metric evaluation, and HFOV--height ambiguity analysis.

    \item \textbf{An evaluation protocol for joint camera--scene consistency.}
    We propose an evaluation protocol with shared scene-level alignment, jointly assessing camera rays, camera poses, rotations, depth, and dense 3D geometry under the same global alignment.
    This protocol prevents independent camera alignment from masking camera--scene inconsistency and is better suited for evaluating the coupled consistency of feed-forward reconstruction outputs.

    \item \textbf{Systematic experimental findings under UAV camera geometry.}
    We systematically evaluate UAVFF3D on multiple representative feed-forward reconstruction models.
    Experiments show that UAV-domain adaptation substantially improves camera-geometry estimation and reconstructed dense scene geometry, 
    mitigates rotation degradation in oblique views, and enhances robustness under HFOV--height variations. 
    We further show that geometric priors are complementary to domain adaptation and improve projection- and alignment-related performance.
\end{itemize}

\section{Related Work}

\subsection{Photogrammetric Reconstruction and Feed-Forward Geometry}
Classical photogrammetric SfM/MVS pipelines remain foundational methods for 3D reconstruction.
These methods explicitly model camera projection, feature matching, bundle adjustment, dense matching, and geometric fusion, and can produce high-quality reconstructions when camera calibration, image overlap, and feature matching are reliable \cite{hartley2003mvg,schonberger2016sfm,schonberger2016mvs}.
Learning-based MVS methods further improve dense reconstruction through cost-volume construction, recurrent regularization, cascaded depth estimation, or PatchMatch-style propagation \cite{yao2018mvsnet,rmvsnet2019,casmvsnet2020,patchmatchnet2021}.
However, these methods usually rely on reliable camera parameters, stable image matching, or explicit multi-view optimization.

Recent feed-forward 3D reconstruction models reformulate the problem by directly predicting geometric outputs in a single network forward pass.
DUSt3R and MASt3R use point-map representations to connect image matching and 3D reconstruction \cite{dust3r2024,mast3r2024}.
VGGT, Pi3, Fast3R, and CUT3R extend feed-forward reconstruction by predicting multi-view camera parameters, depth, point maps, or by supporting efficient and online reconstruction settings
\cite{vggt2025,pi3_2025,fast3r2025,cut3r2025}.
Depth Anything 3 and MapAnything further explore depth--ray representations, metric reconstruction, and optional geometric conditioning or prior-aware inference 
\cite{depthanything3_2025,mapanything2025}.
Their feed-forward design makes them promising for rapid UAV mapping, where camera parameters, depth, and 3D structure must be estimated efficiently
\cite{remondino2011uav_mapping,nex2014uav}.
However, current benchmarks mainly evaluate feed-forward models in general reconstruction scenarios, leaving their camera--scene consistency under UAV-specific camera geometry insufficiently characterized.

\subsection{Camera Ambiguity in Metric 3D Prediction}
The HFOV--height ambiguity studied in this paper is closely related to camera ambiguity in monocular metric depth estimation, where field of view and focal length strongly affect metric scale.
Relative-depth models improve cross-dataset generalization by predicting depth without absolute scale, but it is usually difficult to recover the true focal length, field of view, and metric depth scale from a single RGB image alone \cite{ranftl2020midas}.
Metric monocular depth methods alleviate this limitation through camera-aware normalization, canonical camera transformation, FOV-aware modeling, or direct camera prediction \cite{yin2023metric3d,metric3dv2_2024,piccinelli2024unidepth,sm4depth2024}.
These studies demonstrate that focal length, field of view, and camera intrinsics are essential for metric 3D prediction.

A less explored question is how this camera ambiguity manifests in feed-forward multi-view reconstruction.
Classical multi-view geometry can constrain camera intrinsics and 3D structure through feature correspondences and reprojection consistency \cite{hartley2003mvg,schonberger2016sfm}.
In principle, feed-forward models may also learn similar constraints through attention, point-map consistency, or implicit cross-view matching \cite{dust3r2024,mast3r2024,vggt2025,pi3_2025}.
In practice, however, these models do not guarantee explicit self-calibration or bundle adjustment internally.
When the UAV camera distribution differs from the pretraining distribution, models may rely on learned priors about focal length, scale, or scene layout rather than reliably recovering the true projection geometry from images.
This gives rise to a concrete failure mode in UAV reconstruction: similar image footprints may originate from different HFOV--height configurations, even though they require different camera rays, intrinsics, and metric depth scales.

\subsection{UAV Reconstruction Datasets}

General MVS datasets provide important training and evaluation resources for learning-based 3D reconstruction.
Datasets such as DTU, ETH3D, Tanks and Temples, and BlendedMVS cover diverse indoor and outdoor scenes and are widely used in multi-view geometry, depth estimation, and dense reconstruction research
\cite{dtu2016,eth3d2017,tankstemples2017,yao2020blendedmvs}.
However, these datasets are not designed around UAV photogrammetric acquisition and cannot reflect the UAV camera-geometry variations jointly determined by flight altitude, field of view, viewing direction, and flight pattern.

Existing aerial or UAV-related datasets provide data resources closer to the target scenario.
The WHU-dataset provides aerial-image MVS/Stereo data, WHU-OMVS targets oblique aerial-image reconstruction, and LuoJia-MVS provides five-view aerial-image data
\cite{liu2020whumvs,liu2023whuomvs,luojia_mvs2023}.
ENRICH and UseGeo provide synthetic photogrammetric data and LiDAR-supported real UAV data, respectively
\cite{marelli2023enrich,nex2024usegeo,nex2024usegeo_dataset}.
UAVScenes and UrbanScene3D provide real UAV trajectories, multimodal perception data, or city-scale simulation resources
\cite{uavscenes2025,lin2021urbanscene3d}.
These datasets provide an important basis for UAV reconstruction and urban modeling. However, they are often limited to specific sensors, flight routes, or acquisition patterns, 
and therefore cannot systematically evaluate the robustness of feed-forward reconstruction models under variations in HFOV, flight altitude, nadir/oblique viewing, and controlled HFOV--height ambiguity.

\section{Dataset Design and Evaluation Protocol}

\subsection{UAVFF3D Data Construction Pipeline}
\label{sec:a3d_components_construction}

\begin{figure*}[t]
\centering
\includegraphics[width=0.95\textwidth]{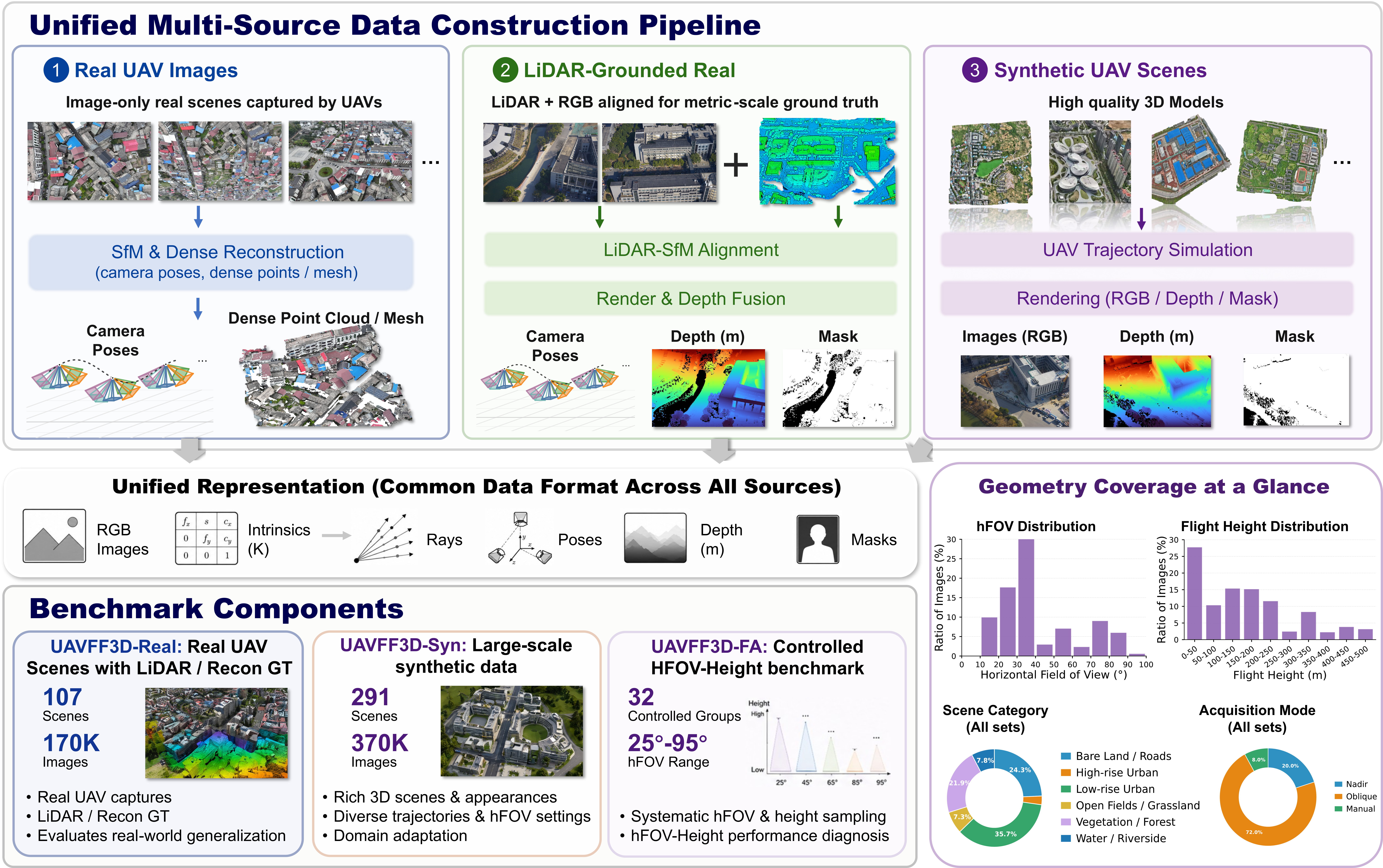}
\caption{
Overview of the UAVFF3D dataset construction pipeline and dataset characteristics.
}
\label{fig:a3d_bench_overview}
\end{figure*}

We aim to construct a geometry-aware benchmark for feed-forward UAV 3D reconstruction.
Unlike datasets that primarily emphasize scene count or appearance diversity, we focus on camera-geometry coverage in UAV acquisition, including HFOV, flight altitude, viewing direction, acquisition pattern, and scene scale.
As shown in Fig.~\ref{fig:a3d_bench_overview}, UAVFF3D adopts a unified multi-source data construction pipeline that converts image-only real UAV data, LiDAR-supported real UAV data, and synthetic UAV scenes into the same data representation.
This unified representation includes RGB images, camera intrinsics, camera rays, camera poses, image-aligned depth maps, and valid masks, enabling data from different sources to share the same interfaces for training, prior-aware inference, and evaluation.

\noindent\textbf{Real UAV Image Branch.}
The first type of input is image-only real UAV imagery.
These data are collected from real UAV acquisition scenarios and capture realistic appearance, flight patterns, and acquisition noise.
For these image blocks, we use a photogrammetric SfM/MVS pipeline to estimate camera poses and reconstruct dense point clouds or meshes.
We then render image-aligned depth maps from the reconstructed geometry and generate valid masks and camera rays.
This branch is mainly used to provide large-scale real UAV appearance and reconstruction-based supervision.

\noindent\textbf{LiDAR-Grounded Real Branch.}
The second type of input consists of LiDAR-supported real UAV data, which are used to build metric-scale reference data under real acquisition conditions.
This branch contains both UAV RGB imagery and independently acquired LiDAR point clouds.
We first perform SfM reconstruction on the RGB imagery to estimate camera poses and establish the SfM image coordinate system.
Following GauU-Scene, which aligns LiDAR point clouds with SfM/COLMAP reconstructions in a common coordinate system, we rigidly register the LiDAR point clouds to the SfM coordinate system and further refine the registration using ICP.
After cross-modal registration, we project the LiDAR point clouds into each camera view and render image-aligned reference depth maps and valid masks.
This branch therefore provides a LiDAR-supported reference for metric reconstruction evaluation in real UAV scenes.

\noindent\textbf{Synthetic UAV Scene Branch.}
The third type of input consists of high-quality textured 3D models.
For synthetic scenes, we simulate UAV acquisition trajectories over the 3D models and explicitly control HFOV, flight altitude, viewing direction, acquisition pattern, and scene scale.
RGB images, depth maps, and valid masks are then obtained by rendering.
This branch is not constrained by real acquisition conditions and can systematically cover camera-geometry configurations that are sparse or missing in real data, such as multiple altitudes, multiple HFOVs, nadir acquisition, oblique acquisition, and irregular local trajectories.
Therefore, the synthetic branch is mainly used to expand camera-geometry coverage in the UAV domain and support UAV-domain adaptation of feed-forward models.

\noindent\textbf{Unified Representation.}
Although the three types of input have different sources and supervision forms, they are ultimately converted into a unified representation.
Each sample contains RGB images, camera intrinsics, camera rays, camera poses, image-aligned depth maps, and valid masks.
This unified format enables real image data, LiDAR-supported data, and synthetic rendered data to share the same interfaces for training, prior-aware inference, and evaluation.

\subsection{Components of UAVFF3D}
\label{sec:a3d_components}

Based on the unified multi-source construction pipeline above, we build the UAVFF3D dataset with three complementary subsets: UAVFF3D-Real, UAVFF3D-Syn, and UAVFF3D-FA.
UAVFF3D-Real connects real UAV appearance, reconstruction supervision, and LiDAR-supported real-world evaluation;
UAVFF3D-Syn provides large-scale controllable synthetic data to expand UAV camera-geometry coverage and support domain adaptation;
UAVFF3D-FA serves as a controlled HFOV--height ambiguity test set for diagnosing feed-forward model behavior when image footprints are approximately the same but projection geometries differ.
Table~\ref{tab:a3d_components} summarizes the scale and geometric settings of the three components.

\begin{table}[h]
\centering
\caption{Overview of the UAVFF3D components.}
\label{tab:a3d_components}
\begingroup
\TableSetup
\begin{tabular}{@{\hspace{0.2em}}lccl@{\hspace{0.2em}}}
\toprule
Component & Source & Scenes & Images \\
\midrule
UAVFF3D-Real & Real & 107 & 170k \\
UAVFF3D-Syn & Synthetic & 291 & 370k \\
UAVFF3D-FA & Synthetic & 32 & 19k \\ 
\bottomrule
\end{tabular}
\endgroup
\end{table}

\noindent\textbf{UAVFF3D-Real.}
UAVFF3D-Real provides real UAV imagery and real-world evaluation data.
It contains 107 processed real scenes and more than 170k images, covering nadir, multi-camera oblique, and manual-flight acquisitions under real flight condition.
For real-world metric evaluation, UAVFF3D-Real further contains three large-scale LiDAR-supported geographic areas.
These LiDAR-supported areas are geographically separated from the UAVFF3D-Real training split.
In each area, we separately acquire five-view oblique and nadir UAV flights and register the LiDAR point cloud to the georeferenced SfM coordinate frame.
The registered LiDAR point cloud is projected and rendered into image views to generate image-aligned reference depth maps and valid masks, providing an independent metric reference for real-world evaluation.
The final evaluation data are divided into 27 processed sub-scenes, including nadir, four-oblique-view, and five-view groups.
Implementation details of LiDAR--SfM registration and reference-depth rendering are provided in Appendix~\ref{sec:appendix_dataset_construction}.

\noindent\textbf{UAVFF3D-Syn.}
UAVFF3D-Syn is large-scale controllable synthetic UAV data for UAV-domain adaptation.
It contains 291 synthetic scenes and more than 370k images rendered from high-quality textured 3D models.
Unlike real UAV data, the camera-geometry configurations in UAVFF3D-Syn are not constrained by original acquisition conditions.
We explicitly control trajectory design, HFOV, flight altitude, viewing direction, scene scale, and acquisition pattern.
Specifically, we design multiple UAV trajectories, including planned nadir mapping routes, planned oblique routes, multi-altitude and multi-HFOV routes, and irregular local trajectories.
These trajectories broaden the UAV camera-geometry distribution seen by the model during fine-tuning and expose feed-forward models to projection and viewpoint configurations that may be sparse or missing in real acquisition.

\begin{figure}[h]
\centering
\includegraphics[width=\linewidth]{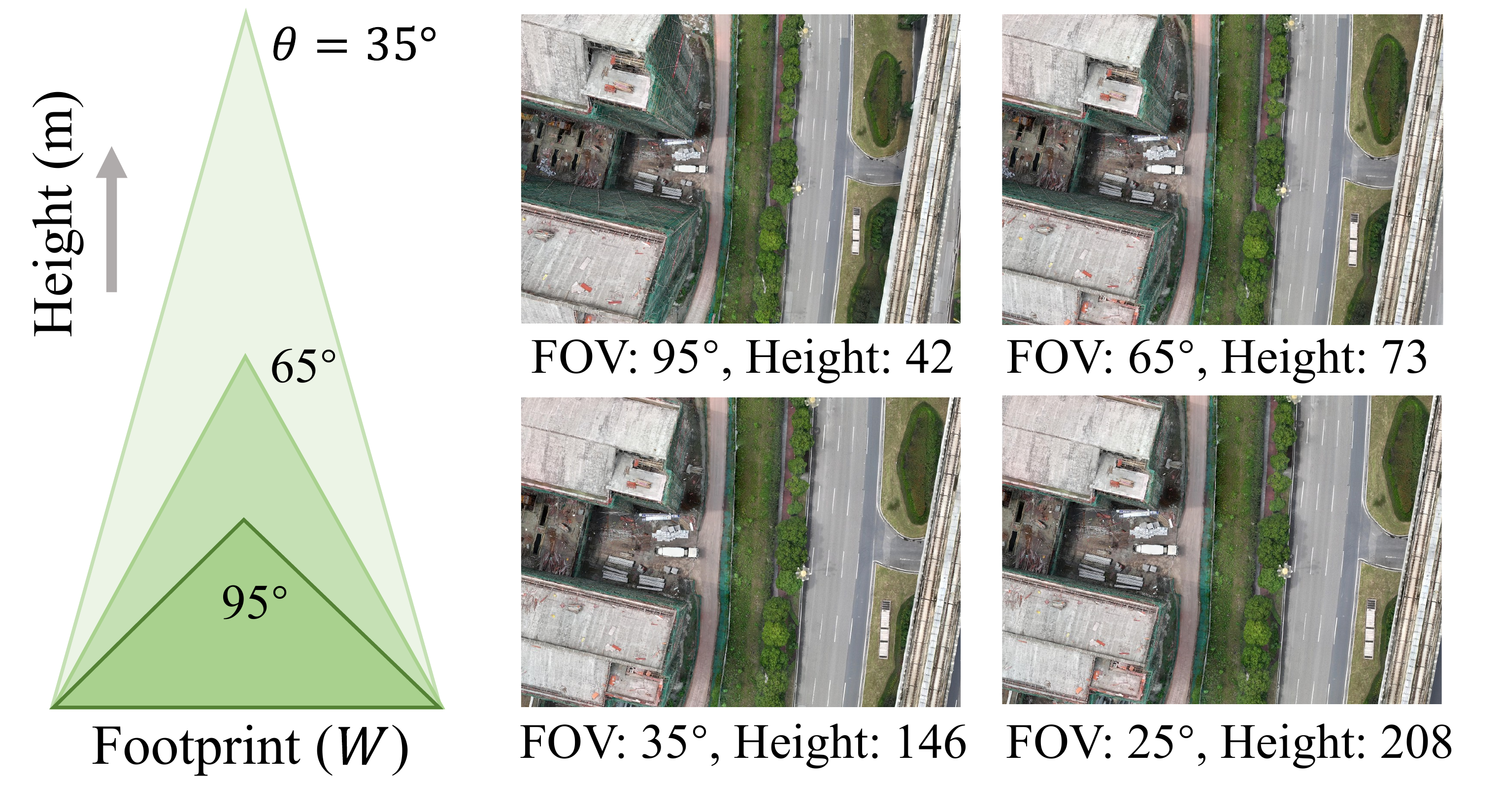}
\caption{
Controlled HFOV--height ambiguity in UAVFF3D-FA.
Different HFOV settings can produce almost identical images in flat regions; 
the resulting images become distinguishable mainly in regions with significant height variation.
}
\label{fig:a3d_fov_design}
\end{figure}

\noindent\textbf{UAVFF3D-FA.}
UAVFF3D-FA is a controlled test set for HFOV--height ambiguity.
It is not intended to increase the scale of synthetic training data, 
but is instead used to evaluate feed-forward models under a specific projection-geometry ambiguity.
The test set contains 32 nadir-view groups generated from four synthetic scenes.
For each scene, we render eight HFOV settings from \(25^\circ\) to \(95^\circ\), and adjust the flight altitude so that the observed ground footprints under different settings remain approximately comparable, as shown in Fig.~\ref{fig:a3d_fov_design}.
Thus, UAVFF3D-FA isolates the effects of HFOV and flight-altitude changes on camera rays, metric scale, and reconstructed geometry under visually similar image content.
Unlike UAVFF3D-Syn, UAVFF3D-FA is excluded from all fine-tuning procedures and is used only as a controlled test set.

The construction of UAVFF3D-FA follows the geometric relationship among HFOV, focal length, flight altitude, and image footprint.
For an image width \(w\) and horizontal field of view \(\theta\), the focal length is
\[
f_x=\frac{w}{2\tan(\theta/2)}.
\]
For a simplified nadir view, the ground-footprint width \(W\) approximately satisfies
\[
W \approx 2H\tan(\theta/2),
\]
where \(H\) is the flight altitude.
Therefore, different HFOV--height pairs can produce similar image footprints, and UAVFF3D-FA uses this relationship to generate controlled groups with comparable footprints but different projection geometries.

\noindent\textbf{Geometry Coverage.}
As shown in Fig.~\ref{fig:a3d_bench_overview}, UAVFF3D provides broad coverage along three dimensions: camera configuration, acquisition pattern, and scene type.
In terms of camera configuration, the dataset contains diverse HFOVs and flight altitudes;
in terms of acquisition pattern, it covers nadir, multi-camera oblique, and manual-flight acquisition;
in terms of scene content, it includes high-rise urban areas, low-rise urban blocks, roads and bare ground, open fields and grassland, vegetation and forests, and water bodies and riverbanks.
This coverage enables UAVFF3D not only to evaluate the overall UAV reconstruction performance of feed-forward models, but also to analyze their robustness under different camera geometries, acquisition patterns, and scene types.

\subsection{Training and Evaluation Data}
\label{sec:data_splits}
\label{sec:evaluation_data_input_settings}

In addition to UAVFF3D-Real and UAVFF3D-Syn, we include BlendedMVS, UAVScenes, and the WHU-dataset for UAV-domain fine-tuning.
BlendedMVS is used to preserve the general multi-view reconstruction capability of the models, while UAVScenes and the WHU-dataset provide additional UAV-domain training data.
All training data are strictly separated from evaluation data, and the UAVFF3D-Real test split and UAVFF3D-FA are not involved in any fine-tuning procedure.

For evaluation, we use four UAV test datasets: UseGeo~\cite{nex2024usegeo}, UrbanScene3D~\cite{lin2021urbanscene3d}, the UAVFF3D-Real test split, and UAVFF3D-FA.
UseGeo, UrbanScene3D, and UAVFF3D-Real support the main transfer evaluation, while UAVFF3D-FA provides a controlled HFOV--height diagnostic setting.
The main transfer evaluation covers nadir, four-oblique-view, and five-view acquisition, with HFOV ranging from \(36^\circ\) to \(81^\circ\) and flight altitude ranging from 80 to 191~m.
UAVFF3D-FA further provides a controlled diagnostic setting, extending the HFOV range to \(25^\circ\)--\(95^\circ\) and the flight-altitude range to approximately 40--210~m.
Detailed HFOV and altitude settings of each dataset are given in Appendix~\ref{sec:appendix_dataset_processing}.
This training--evaluation separation allows the benchmark to test both real-world transfer capability and controlled projection-geometry sensitivity while avoiding data overlap.
Table~\ref{tab:data_splits} summarizes the training and evaluation data.

\begin{table}[h]
\centering
\caption{
Datasets used for UAV-domain fine-tuning and evaluation.
The first five datasets are used for UAV-domain fine-tuning, whereas the last four datasets are used only for evaluation.
}
\label{tab:data_splits}
\begingroup
\TableSetup
\begin{tabular}{@{\hspace{0.4em}}lcc@{\hspace{0.4em}}}
\toprule
Dataset & Acquisition & Images \\
\midrule
BlendedMVS & generic & 226k \\
UAVScenes & nadir & 24k \\
WHU-dataset & nadir & 1.7k \\
UAVFF3D-Real(train) & nadir/oblique & 85k \\
UAVFF3D-Syn & nadir/oblique/manual & 370k \\
\midrule
UseGeo & nadir & 828 \\
UrbanScene3D & oblique & 12k \\
UAVFF3D-Real(test) & nadir/oblique & 85k \\
UAVFF3D-FA & nadir & 19k \\
\bottomrule
\end{tabular}
\endgroup
\end{table}

\subsection{Evaluation Protocol}
\label{sec:evaluation_protocol}

\begin{figure}[h]
\centering
\includegraphics[width=\linewidth]{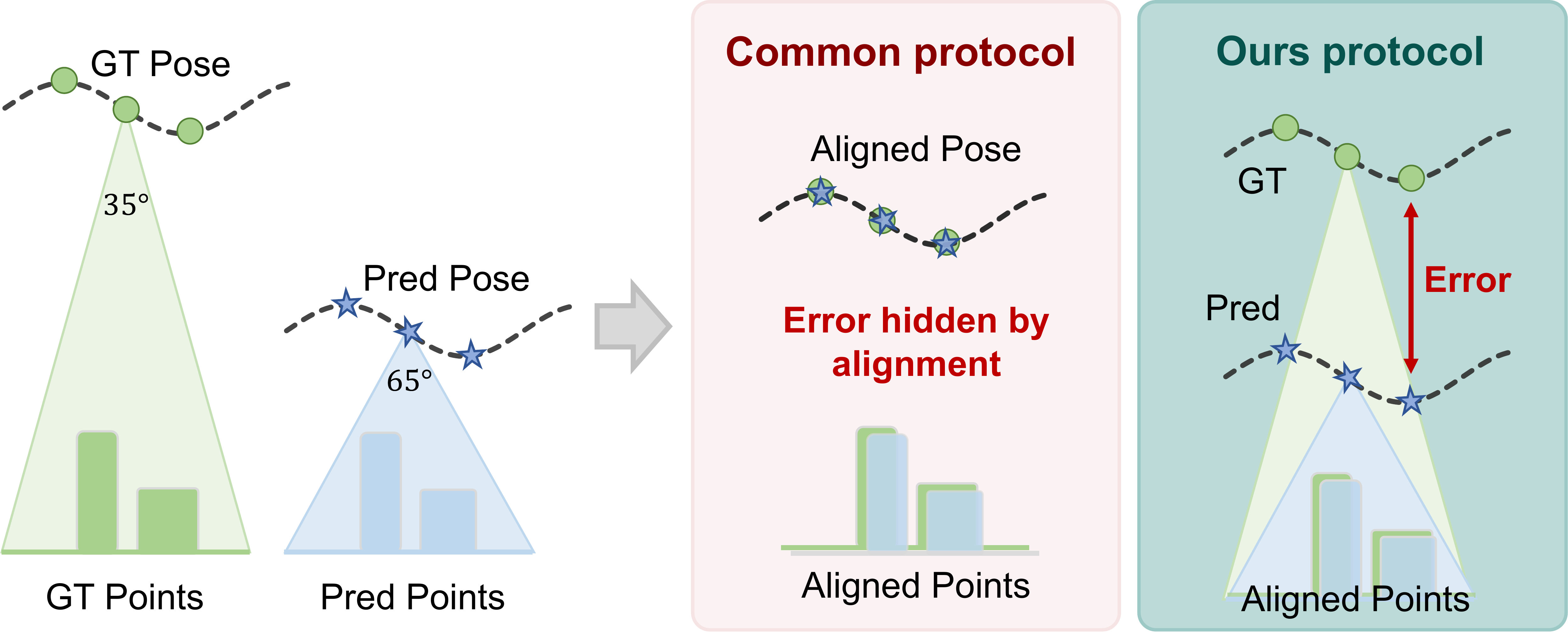}
\caption{
Comparison between the commonly used separate-alignment protocol and the shared-alignment protocol of UAVFF3D.
}
\label{fig:evaluation_protocol}
\end{figure}

Current feed-forward reconstruction evaluations usually align dense geometry and camera poses separately before computing the corresponding metrics.
As shown in Fig.~\ref{fig:evaluation_protocol}, the predicted point cloud can be aligned to the GT point cloud for geometric evaluation, while the predicted camera trajectory is independently aligned to the GT trajectory for pose evaluation.
Although this protocol may report low geometry and pose errors after two independent alignments, it ignores an important property of feed-forward reconstruction outputs: the predicted cameras and reconstructed geometry are defined in the same coordinate system and share the same scale.
Therefore, separate alignment may mask camera--scene inconsistency because the predicted geometry and predicted cameras are no longer evaluated as a coupled reconstruction result.

\begin{figure*}[t]
\centering
\includegraphics[width=0.95\textwidth]{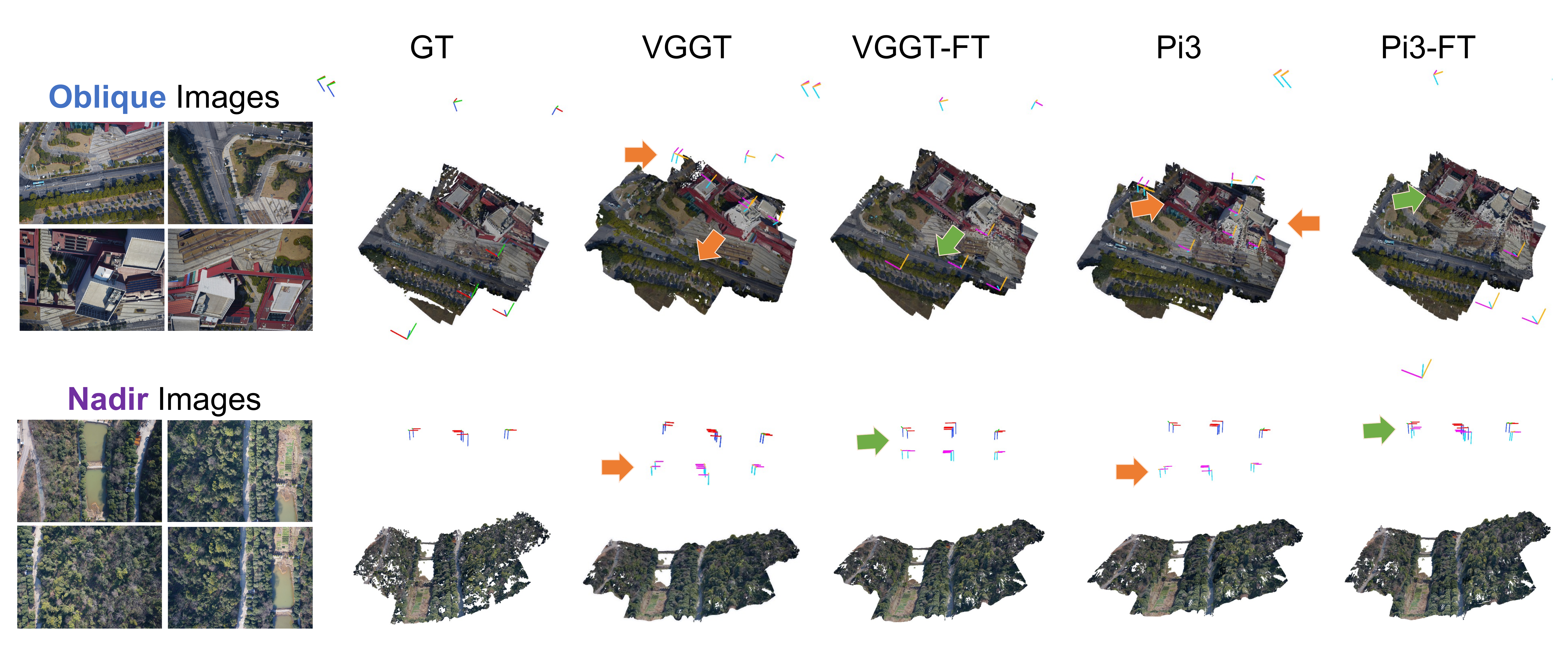}
\caption{Qualitative effect of UAVFF3D fine-tuning. First row: oblique input images, where fine-tuning improves pose and point-cloud consistency. Second row: nadir input images, where fine-tuning improves camera and point-cloud consistency.}
\label{fig:qualitative_ft}
\label{fig:vggt_ft_vis}
\end{figure*}

We therefore evaluate the predicted cameras and dense geometry as a coupled reconstruction result.
Let the predicted dense point set be \(\hat{\mathcal{X}}\), the GT point set be \(\mathcal{X}\), and the predicted camera poses be \(\{\hat{\mathbf{T}}_i\}_{i=1}^{N}\).
For each predicted reconstruction, we estimate only one scene-level global similarity transform \(\mathbf{S}^{\star}\in \mathrm{Sim}(3)\), such that the aligned predicted point set is as consistent as possible with the GT point set in the scene coordinate system.
The same \(\mathbf{S}^{\star}\) is then consistently applied to the dense points and all predicted camera poses, i.e., we jointly evaluate \(\mathbf{S}^{\star}\hat{\mathcal{X}}\) and \(\mathbf{S}^{\star}\hat{\mathbf{T}}_i\).
This shared alignment can be estimated from either the predicted point cloud or the predicted camera trajectory.
We use point-cloud-based alignment because dense geometry provides many more spatial observations than a small number of camera centers and is therefore more robust for scene-level metric registration.
Under this setting, the pose metric no longer measures whether the predicted camera trajectory can be independently transformed to fit the GT trajectory; instead, it measures whether the predicted cameras remain consistent with the reconstructed geometry after the shared scene-level alignment.

We report five complementary metrics to measure different aspects of reconstruction quality.
Let the predicted camera center, rotation, depth, and camera ray be
\(\hat{\mathbf{c}}_i, \hat{\mathbf{R}}_i, \hat{d}_{i,p}, \hat{\mathbf{r}}_{i,p}\),
and the corresponding GT quantities be \(\mathbf{c}_i, \mathbf{R}_i, d_{i,p}, \mathbf{r}_{i,p}\).
Further denote
\(\hat{\mathcal{X}}^{\star}=\mathbf{S}^{\star}\hat{\mathcal{X}}\),
\(\hat{\mathbf{c}}_i^{\star}=\mathbf{S}^{\star}\hat{\mathbf{c}}_i\),
and let \(\hat{d}_{i,p}^{\,\star}\) be the predicted depth after applying the same scene-level alignment.
We define the one-way Chamfer-L1 distance as
\(D(\mathcal{A},\mathcal{B})=\mathrm{mean}_{\mathbf{a}\in\mathcal{A}}
\min_{\mathbf{b}\in\mathcal{B}}\|\mathbf{a}-\mathbf{b}\|_1\).
Based on this notation, the metrics are defined as
\[
\begin{aligned}
E_{\mathrm{ray}} &= \mathrm{mean}_{i,p}\,\angle(\hat{\mathbf{r}}_{i,p},\mathbf{r}_{i,p}), \\
E_{\mathrm{ATE}} &= \mathrm{mean}_{i}\,\|\hat{\mathbf{c}}_i^{\star}-\mathbf{c}_i\|_2, \\
E_{\mathrm{AbsRel}} &= \mathrm{mean}_{i,p}\,
\frac{|\hat{d}_{i,p}^{\,\star}-d_{i,p}|}{d_{i,p}}, \\
E_{\mathrm{rot}} &= \mathrm{mean}_{i}\,\angle(\hat{\mathbf{R}}_i,\mathbf{R}_i), \\
E_{\mathrm{CD}} &= \frac{1}{2}
\bigl(D(\hat{\mathcal{X}}^{\star},\mathcal{X})
+ D(\mathcal{X},\hat{\mathcal{X}}^{\star})\bigr).
\end{aligned}
\]
Here, Ray Error measures the projection-ray error, Pose ATE measures the camera-center error after shared alignment, AbsRel Depth measures local depth error, Rotation MAE measures the camera-orientation error, and Chamfer-L1 measures dense geometric error.
Before computing Chamfer-L1 distance, both the aligned predicted point cloud and the GT point cloud are downsampled with a voxel size of \(0.25~\mathrm{m}\).

\section{Experiments}
\label{sec:experiments}

\subsection{Experimental Setup}
\label{sec:experimental_setup}

We evaluate five representative feed-forward reconstruction models: VGGT, Pi3, MapAnything, Pi3X, and Depth Anything 3 (DA3)~\cite{vggt2025,pi3_2025,mapanything2025,depthanything3_2025}.
All models are evaluated with RGB-only inputs to compare their UAV-domain generalization capability.
For MapAnything, Pi3X, and DA3, which support explicit geometric priors, we further evaluate the C, P, and CP input settings, where C denotes camera intrinsics, P denotes camera poses, and CP denotes providing both.

All zero-shot experiments use the released pretrained checkpoints.
For UAV-domain adaptation, we fine-tune VGGT, Pi3, MapAnything, and Pi3X using the training data defined in Section~\ref{sec:data_splits}.
Because the training code of DA3 is not publicly available, DA3 is used only as a zero-shot baseline in this paper.
During training, we use a fixed sampling mixture and sample from BlendedMVS, UAVFF3D-Real, UAVFF3D-Syn, UAVScenes, and the WHU-dataset with a ratio of \(20{:}20{:}40{:}1{:}1\), respectively.
This setting preserves general multi-view reconstruction capability while increasing the proportion of controllable UAV camera geometry from UAVFF3D-Syn and avoiding over-sampling of small public UAV datasets during training.
All reported metrics follow the shared-alignment evaluation protocol in Section~\ref{sec:evaluation_protocol}.
For each model, dataset, and input setting, metrics are first averaged over sampled multi-view image sets with the same number of input views, and the final score is then averaged over the 8-, 16-, 24-, and 32-view settings.

All fine-tuning experiments are implemented within the MapAnything training framework to ensure a unified data interface, sampling strategy, and training pipeline across different models.
Models are trained with variable-sized multi-view image sets containing 2 to 8 views.
Input images are resized to 518 pixels, and random multi-resolution scaling augmentation is further applied during training.
Training uses the AdamW optimizer with an initial learning rate of \(5\times10^{-6}\), a minimum learning rate of \(5\times10^{-8}\), a 2-epoch warm-up, and half-cycle cosine decay.
All models are fine-tuned for 10 epochs on two NVIDIA RTX A6000 GPUs or two NVIDIA A100 40GB GPUs.

\subsection{Necessity of the Evaluation Protocol}
\label{sec:alignment_protocol_main_ablation}

Before analyzing UAV-domain adaptation, we first validate the necessity of the shared-alignment evaluation protocol.
Table~\ref{tab:alignment_protocol_main} compares pose errors under shared alignment and separate camera alignment.
CD-S and ATE-S are computed using the same scene-level alignment estimated from the point cloud, while ATE-I is computed using independent camera-center alignment.
The Gap indicates the difference between the two ATE values, namely the camera--scene inconsistency removed by independent camera alignment.

\begin{table}[h]
\centering
\caption{Comparison between shared alignment and separate camera alignment. All results use zero-shot VGGT.}
\label{tab:alignment_protocol_main}
\begingroup
\TableSetup
\begin{tabular}{@{\hspace{0.4em}}lcccc@{\hspace{0.4em}}}
\toprule
Dataset & CD-S$\downarrow$ & ATE-S$\downarrow$ & ATE-I$\downarrow$ & Gap$\downarrow$ \\
\midrule
UseGeo & 1.61 & 8.32 & 3.36 & 4.96 \\
UrbanScene3D & 4.75 & 77.78 & 44.51 & 33.27 \\
UAVFF3D-Real & 3.38 & 89.45 & 47.21 & 42.24 \\
UAVFF3D-FA & 1.04 & 41.62 & 1.11 & 40.52 \\
\midrule
Average & 2.70 & 54.29 & 24.05 & 30.25 \\
\bottomrule
\end{tabular}
\endgroup
\end{table}

The results show that, on all datasets, ATE-I obtained by independent camera alignment is consistently much lower than ATE-S under shared alignment.
On UrbanScene3D and UAVFF3D-Real, Pose ATE remains large even after separate camera alignment, suggesting that oblique imagery leads to substantial pose-estimation errors.
On UAVFF3D-FA, ATE-I is only 1.11, whereas ATE-S reaches 41.62, indicating that the predicted camera trajectory can be fitted to the GT trajectory in isolation, but it is not in a scene coordinate system consistent with the predicted dense geometry.
Therefore, separate alignment yields overly optimistic pose errors and fails to reflect the coupled consistency between cameras and scene geometry in feed-forward reconstruction.
This validates the necessity of using shared scene-level alignment for joint evaluation.

\subsection{Effect of Fine-Tuning on UAV Reconstruction}
\label{sec:a3d_finetuning_effect}
\label{sec:effect_of_uav_domain_adaptation}

\begin{figure}[h]
\centering
\includegraphics[width=\linewidth]{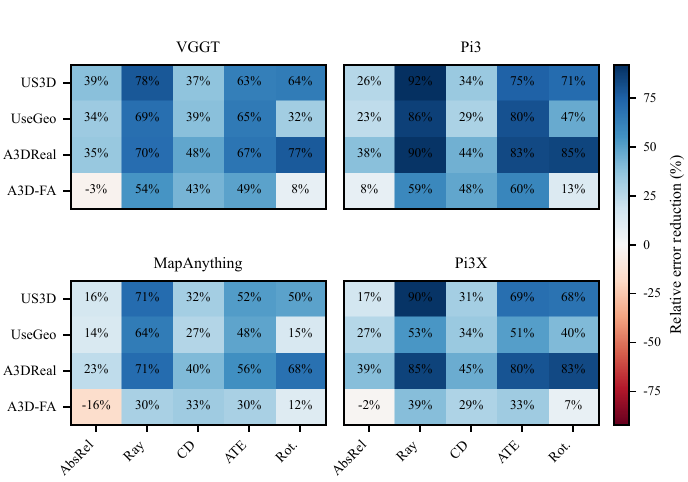}
\caption{
Dataset-level effect of fine-tuning.
Each cell reports the relative error reduction; positive values indicate improvement after fine-tuning.
Red denotes error reduction, and blue denotes improvement.
}
\label{fig:fine_tuning_improvement_heatmap}
\end{figure}

Table~\ref{tab:main_adaptation} compares the pretrained models and their fine-tuned versions.
We report aggregated results over four UAV test datasets and visualize dataset-level changes in Fig.~\ref{fig:fine_tuning_improvement_heatmap}.
For each dataset--metric pair, the relative error reduction is defined as
\(\Delta=(e_{\mathrm{pre}}-e_{\mathrm{FT}})/e_{\mathrm{pre}}\times100\%\), where \(e_{\mathrm{pre}}\) and \(e_{\mathrm{FT}}\) denote the errors of the pretrained and fine-tuned models, respectively.
A positive \(\Delta\) indicates an improvement after fine-tuning.

\begin{table}[h]
\centering
\caption{Overall UAV reconstruction performance before and after fine-tuning, averaged over four UAV test datasets.}
\label{tab:main_adaptation}
\label{tab:adapt_main}
\begingroup
\TableSetup
\begin{tabular}{@{\hspace{0.4em}}lccccc@{\hspace{0.4em}}}
\toprule
Model & AbsRel$\downarrow$ & Ray$\downarrow$ & CD$\downarrow$ & ATE$\downarrow$ & Rot.$\downarrow$ \\
\midrule
VGGT & 0.042 & 6.23 & 2.70 & 54.29 & 11.60 \\
Pi3 & 0.036 & 8.67 & 2.86 & 60.39 & 11.41 \\
MapAnything & 0.044 & 6.00 & 2.90 & 57.50 & 13.33 \\
Pi3X & 0.035 & 7.73 & 2.24 & 56.19 & 11.72 \\
DA3 & 0.037 & 6.58 & 2.94 & 56.21 & 14.79 \\
\midrule
VGGT-FT & 0.028 & 1.97 & 1.59 & 20.57 & 3.96 \\
Pi3-FT & \textbf{0.026} & \textbf{1.37} & 1.77 & \textbf{14.52} & \textbf{3.07} \\
Mapa-FT & 0.037 & 2.51 & 1.91 & 29.08 & 6.10 \\
Pi3X-FT & 0.026 & 1.87 & \textbf{1.44} & 18.70 & 3.40 \\
\bottomrule
\end{tabular}
\endgroup
\end{table}

Table~\ref{tab:main_adaptation} shows that fine-tuning consistently improves geometry-sensitive metrics.
The largest gains occur in Ray Error, Pose ATE, and Rotation MAE, indicating better consistency among predicted camera rays, camera poses, scene scale, and dense geometry.
The dataset-level heatmap in Fig.~\ref{fig:fine_tuning_improvement_heatmap} further shows that the improvement is not dominated by a single test source.
Together with the qualitative comparison in Fig.~\ref{fig:qualitative_ft}, these results demonstrate that UAVFF3D-based domain adaptation provides broad robustness gains under UAV acquisition geometries.

Table~\ref{tab:data_ablation} disentangles the effects of the synthetic dataset and real UAV data.
Adding UAVFF3D-Syn improves camera-geometry metrics, showing the value of explicitly covering UAV HFOVs, flight altitudes, and trajectory patterns.
UAVFF3D-Real further provides real appearance statistics, acquisition irregularities, and scene-layout characteristics.
Combining the two yields the most balanced performance, indicating that synthetic camera-geometry coverage and real UAV data are complementary.

\begin{table}[h]
\centering
\caption{
Training-data ablation for UAV-domain adaptation.
Public denotes fine-tuning only on public training datasets;
Real denotes adding UAVFF3D-Real training data;
Syn denotes adding UAVFF3D-Syn data;
Full uses all data.
}
\label{tab:data_ablation}
\begingroup
\TableSetup
\begin{tabular}{@{\hspace{0.4em}}lccccc@{\hspace{0.4em}}}
\toprule
Model & AbsRel$\downarrow$ & Ray$\downarrow$ & CD$\downarrow$ & ATE$\downarrow$ & Rot.$\downarrow$ \\
\midrule
Pi3 & 0.036 & 8.67 & 2.86 & 60.39 & 11.41 \\
Pi3-FT-Public & 0.032 & 5.59 & 1.81 & 46.74 & 8.76 \\
Pi3-FT-Syn & 0.032 & 2.36 & 1.65 & 25.80 & 5.31 \\
Pi3-FT-Real & 0.023 & 1.73 & 2.15 & 16.55 & 2.97 \\
Pi3-FT-Full & 0.026 & 1.37 & 1.77 & 14.52 & 3.07 \\
\midrule
MapAnything & 0.044 & 6.00 & 2.90 & 57.50 & 13.33 \\
Mapa-FT-Public & 0.044 & 5.39 & 3.33 & 55.54 & 13.21 \\
Mapa-FT-Syn & 0.052 & 3.28 & 2.64 & 51.10 & 11.94 \\
Mapa-FT-Real & 0.041 & 3.16 & 2.45 & 30.55 & 6.25 \\
Mapa-FT-Full & 0.037 & 2.51 & 1.91 & 29.08 & 6.10 \\
\bottomrule
\end{tabular}
\endgroup
\end{table}

\begin{figure*}[t]
\centering
\includegraphics[width=0.95\textwidth]{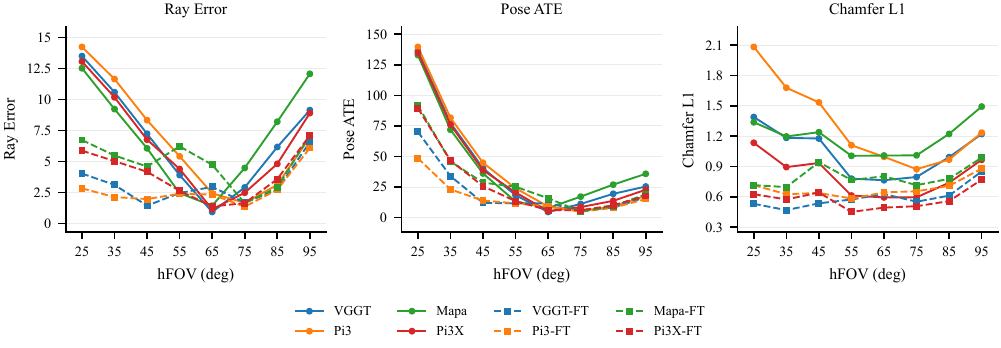}
\caption{
Controlled HFOV--height diagnosis on UAVFF3D-FA.
}
\label{fig:HFOV_diagnosis}
\end{figure*}

\subsection{Camera-Geometry Analysis under UAV Acquisition}
\label{sec:camera_geometry_analysis}
\label{sec:oblique_view_degradation}
\label{sec:HFOV_controlled_diagnosis}

In this section, we examine whether the benefits of fine-tuning are consistent across different UAV camera geometries.
We focus on two stress scenarios: oblique acquisition and controlled HFOV--height variation.
The former tests model robustness to viewpoint-dependent visibility and depth-distribution changes, while the latter tests whether models can handle different projection geometries when image footprints are approximately comparable.

\paragraph{Oblique-view geometry.}
Table~\ref{tab:oblique_gap_main} compares oblique and nadir reconstruction before and after fine-tuning.
The main table reports CD and Rotation MAE, as they summarize dense scene consistency and camera-orientation errors, respectively.
For each lower-is-better metric, we define the oblique--nadir gap as
\(\mathrm{Gap}=e_{\mathrm{oblique}}-e_{\mathrm{nadir}}\).
A positive gap means that oblique views remain more difficult than near-nadir views, while a smaller gap indicates more balanced performance across acquisition modes.
The complete oblique--nadir breakdown including prior-aware input modes is provided in Appendix Table~\ref{tab:appendix_oblique_nadir_prior_results}.

\begin{table}[h]
\centering
\caption{Oblique and nadir reconstruction before and after fine-tuning.}
\label{tab:oblique_gap_main}
\begingroup
\TableSetup
\begin{tabular}{@{\hspace{0.4em}}lcccccc@{\hspace{0.4em}}}
\toprule
\multirow{2}{*}{Model} 
& \multicolumn{3}{c}{CD$\downarrow$}
& \multicolumn{3}{c}{Rot.$\downarrow$} \\
\cmidrule(lr){2-4}\cmidrule(lr){5-7}
& Oblique & Nadir & Gap & Oblique & Nadir & Gap \\
\midrule
VGGT & 5.42 & 2.02 & 3.40 & 31.89 & 6.32 & 25.57 \\
MapAnything & 6.51 & 1.79 & 4.72 & 39.02 & 5.12 & 33.90 \\
Pi3 & 5.80 & 2.29 & 3.50 & 33.00 & 6.86 & 26.13 \\
Pi3X & 5.60 & 1.89 & 3.71 & 35.89 & 10.11 & 25.78 \\
DA3 & 5.84 & 2.05 & 3.79 & 32.95 & 10.81 & 22.14 \\
\midrule
VGGT-FT & 2.86 & 1.19 & 1.68 & 7.79 & 3.07 & 4.73 \\
Pi3-FT & 2.89 & 1.49 & \textbf{1.40} & \textbf{5.31} & \textbf{2.88} & \textbf{2.43} \\
Mapa-FT & 3.67 & 1.23 & 2.44 & 13.06 & 3.64 & 9.43 \\
Pi3X-FT & \textbf{2.53} & \textbf{1.10} & 1.43 & 6.29 & 2.95 & 3.35 \\
\bottomrule
\end{tabular}
\endgroup
\end{table}

Pretrained models exhibit a large oblique--nadir gap, especially in Rotation MAE, indicating that oblique acquisition primarily degrades camera orientation estimation and dense scene consistency.
After fine-tuning, both oblique and nadir reconstruction improve, and more importantly, the gap between them is greatly reduced.
For example, the rotation gap of VGGT decreases from 25.57 to 4.73, and that of MapAnything decreases from 33.90 to 9.43.

\paragraph{HFOV--height projection geometry.}
UAVFF3D-FA evaluates projection-geometry sensitivity by varying HFOV and flight altitude while keeping the observed footprint approximately comparable.
This setting tests whether feed-forward models can recover the correct projection geometry when it cannot be fully disambiguated from image content alone.

Figure~\ref{fig:HFOV_diagnosis} shows that RGB-only inference exhibits a clear dependence on HFOV on UAVFF3D-FA.
The Ray Error and Pose ATE of pretrained models are usually lower around \(65^\circ\)--\(75^\circ\), but increase markedly under narrower or wider HFOVs. This suggests that RGB-only feed-forward models are strongly influenced by implicit camera-geometry priors in the training distribution.
After fine-tuning, all models improve consistently across different HFOV settings, with overall reductions in Ray Error, Pose ATE, and Chamfer-L1.
At the same time, the error curves become flatter, showing that UAVFF3D domain adaptation can reduce the models' dependence on a particular HFOV range and improve robustness to HFOV--height variations.
However, the Ray Error and Pose ATE of fine-tuned models remain high at extreme HFOVs, indicating that recovering camera rays, focal-length-related projection geometry, and the camera--scene scale relationship from RGB images alone remains difficult.

\subsection{Explicit Geometric Priors}
\label{sec:prior_adaptation_interaction}
\label{sec:effect_of_geometric_priors}

Beyond RGB-only inference, some feed-forward reconstruction models support external geometric information such as camera intrinsics or poses at inference time.
Because such information is often available in UAV photogrammetry through flight platforms, camera calibration, or photogrammetric processing, this section analyzes the relationship between explicit geometric priors and UAV-domain adaptation.

\begin{table}[h]
\centering
\caption{
Interaction between explicit geometric priors and UAV-domain adaptation.
The table compares RGB-only and prior-aware inputs before and after UAV-domain adaptation.
C, P, and CP denote providing camera intrinsics, camera poses, and their combination, respectively.
Blue/red cells indicate improvement/degradation relative to the RGB input of the same model.
All metrics are lower-is-better.
}
\label{tab:prior_adaptation}
\label{tab:prior_main}
\begingroup
\TableSetup
\begin{tabular}{@{\hspace{0.4em}}lcccccc@{\hspace{0.4em}}}
\toprule
Model & Input & AbsRel$\downarrow$ & Ray$\downarrow$ & CD$\downarrow$ & ATE$\downarrow$ & Rot.$\downarrow$ \\
\midrule
MapAnything & RGB & 0.044 & 6.00 & 2.90 & 57.50 & 13.33 \\
MapAnything & C & \cellcolor{blue!1}0.043 & \cellcolor{blue!39}0.96 & \cellcolor{blue!6}2.51 & \cellcolor{blue!15}38.52 & \cellcolor{blue!3}12.32 \\
MapAnything & P & \cellcolor{red!55}0.105 & \cellcolor{blue!2}5.79 & \cellcolor{red!44}5.66 & \cellcolor{red!5}63.97 & \cellcolor{blue!21}7.34 \\
MapAnything & CP & \cellcolor{red!55}0.099 & \cellcolor{blue!41}0.66 & \cellcolor{red!36}5.15 & \cellcolor{blue!10}44.72 & \cellcolor{blue!19}7.73 \\
Pi3X & RGB & 0.035 & 7.73 & 2.24 & 56.19 & 11.72 \\
Pi3X & C & \cellcolor{blue!3}0.032 & \cellcolor{blue!44}0.24 & \cellcolor{blue!3}2.08 & \cellcolor{blue!27}23.17 & \cellcolor{blue!10}9.27 \\
Pi3X & P & \cellcolor{blue!4}0.032 & \cellcolor{blue!2}7.44 & \cellcolor{blue!2}2.14 & \cellcolor{blue!6}48.23 & \cellcolor{blue!21}6.46 \\
Pi3X & CP & \cellcolor{blue!7}0.029 & \cellcolor{blue!44}\textbf{0.23} & \cellcolor{blue!4}2.04 & \cellcolor{blue!35}13.48 & \cellcolor{blue!29}4.38 \\
DA3 & RGB & 0.037 & 6.58 & 2.94 & 56.21 & 14.79 \\
DA3 & CP & \cellcolor{red!3}0.039 & \cellcolor{blue!28}2.51 & \cellcolor{red!3}3.11 & \cellcolor{blue!23}27.88 & \cellcolor{blue!9}11.86 \\
\midrule
\multicolumn{7}{l}{\textit{Prior-aware evaluation after UAV-domain adaptation}} \\
\midrule
Mapa-FT & RGB & 0.037 & 2.51 & 1.91 & 29.08 & 6.10 \\
Mapa-FT & C & \cellcolor{blue!5}0.033 & \cellcolor{blue!39}0.38 & \cellcolor{blue!1}1.86 & \cellcolor{blue!21}15.86 & \cellcolor{blue!4}5.55 \\
Mapa-FT & P & \cellcolor{blue!15}0.025 & \cellcolor{blue!2}2.38 & \cellcolor{blue!14}1.35 & \cellcolor{blue!18}17.61 & \cellcolor{blue!34}1.61 \\
Mapa-FT & CP & \cellcolor{blue!20}0.021 & \cellcolor{blue!40}0.30 & \cellcolor{blue!19}\textbf{1.14} & \cellcolor{blue!40}\textbf{3.43} & \cellcolor{blue!38}\textbf{1.06} \\
Pi3X-FT & RGB & 0.026 & 1.87 & 1.44 & 18.70 & 3.40 \\
Pi3X-FT & C & \cellcolor{blue!10}0.020 & \cellcolor{blue!40}0.25 & \cellcolor{red!6}1.63 & \cellcolor{blue!26}8.13 & \cellcolor{blue!4}3.14 \\
Pi3X-FT & P & \cellcolor{blue!9}0.021 & \cellcolor{red!1}1.93 & \cellcolor{red!11}1.77 & \cellcolor{blue!5}16.50 & \cellcolor{blue!19}1.98 \\
Pi3X-FT & CP & \cellcolor{blue!15}\textbf{0.018} & \cellcolor{blue!39}0.26 & \cellcolor{red!14}1.89 & \cellcolor{blue!34}4.80 & \cellcolor{blue!24}1.62 \\
\bottomrule
\end{tabular}
\endgroup
\end{table}

In the zero-shot setting, the benefits of explicit geometric priors are clearly metric-dependent.
Camera intrinsics usually reduce Ray Error substantially, which is consistent with their direct constraint on projection rays.
Some prior settings also improve Pose ATE or Rotation MAE, suggesting that external camera information can partially reduce projection- and camera-alignment-related errors.
However, such improvement does not always transfer to all geometric metrics.
For example, the AbsRel and Chamfer-L1 of MapAnything degrade markedly under P and CP inputs, and the CP input of DA3 does not yield consistent dense-geometry improvement.
This suggests that, for models not adapted to the UAV domain, geometric priors introduced at test time may not fully match the pretrained assumptions about scale, focal length, pose layout, and scene geometry, resulting in trade-offs among different metrics.

After UAV-domain adaptation, the effects of prior inputs become more stable overall.
For Mapa-FT, the C, P, and CP settings improve almost all metrics relative to RGB-only inference, with the CP combination achieving the strongest overall performance.
Pi3X-FT also clearly benefits from camera intrinsics, especially in Ray Error and Pose ATE.
However, its CP setting degrades Chamfer-L1, indicating that explicit priors do not guarantee monotonic improvement across all metrics.

\section{Conclusion}
This paper addresses camera--scene consistency in feed-forward UAV 3D reconstruction and proposes UAVFF3D, a geometry-aware real--synthetic benchmark.
Rather than attributing model failure solely to appearance-domain shift, our experiments show that variations in viewing angle, field of view, flight altitude, and acquisition pattern in UAV imagery significantly affect the joint prediction of camera rays, poses, scale, and reconstructed dense scene geometry by feed-forward models.
Therefore, the core challenge of feed-forward UAV reconstruction lies not only in generating locally plausible depth maps or point clouds, but also in maintaining consistency between camera geometry and scene structure in a unified coordinate system.

Our analysis on UAVFF3D further reveals two representative geometric failure modes.
Oblique acquisition introduces stronger perspective changes, occlusion effects, and depth-distribution variation, which primarily disrupt rotation estimation and dense scene consistency.
The HFOV--height analysis shows that, when image footprints are similar, RGB-only feed-forward models may still struggle to distinguish different projection geometries and metric scales.
These phenomena indicate that model instability in the UAV domain is not an isolated degradation on a single metric or dataset, but is closely related to a mismatch between implicit camera priors and the UAV camera-geometry distribution.

The experimental results show that UAV-domain adaptation improves model robustness to UAV camera-geometry distributions and enhances the overall consistency of coupled camera--scene reconstruction.
Synthetic data provide controllable and broadly covered camera-geometry variations, while real data complement them with appearance, trajectory, and scene-distribution characteristics from practical acquisition; the two are complementary.
Meanwhile, explicit camera priors directly constrain projection- and alignment-related variables and yield more stable gains after UAV-domain adaptation.
This suggests that reliable feed-forward UAV reconstruction cannot rely only on larger-scale image training data, but must also consider camera-geometry coverage, evaluation-protocol design, and the appropriate use of prior information.

Nevertheless, results on UAVFF3D show that current models have not fully resolved the projection-geometry ambiguity caused by HFOV--height variations. This limitation is especially evident under extreme fields of view or in the absence of external camera information, where substantial errors remain.
Future work should further explore how to resolve this ambiguity to improve model stability and generalization under complex UAV camera-geometry conditions.

{\begin{spacing}{1.17}
\normalsize
\bibliography{refs}
\end{spacing}}

\FloatBarrier
\clearpage

\appendix
\clearpage

\section{UAVFF3D Data Construction Details}
\label{sec:appendix_dataset_construction}
This appendix provides additional details on UAVFF3D data construction and the complete experimental results.
To avoid repetition with the main paper, the appendix includes only processing procedures, camera-geometry settings, and supplementary quantitative results that are not discussed in detail in the main text.

\subsection{Real UAV and LiDAR-Supported Test Data}
\label{sec:appendix_real_lidar}
\label{sec:appendix_uav_platform}
The LiDAR-supported UAVFF3D-Real test data were collected using a DJI Matrice 350 RTK platform.
Five-view oblique images were acquired using a PSDK 102S V2 oblique imaging payload, nadir images were acquired from separate downward-looking flights, and LiDAR point clouds were acquired using a DJI Zenmuse L2.
The three test areas are used only for evaluation and are geographically separated from the UAVFF3D-Real training areas.

\begin{table}[h]
\centering
\caption{LiDAR-supported UAVFF3D-Real test acquisition areas and metadata. These areas are geographically separated from the UAVFF3D-Real training split.}
\label{tab:a3d_real_test}
\begingroup
\TableSetup
\begin{tabular}{@{\hspace{0.4em}}ccccc@{\hspace{0.4em}}}
\toprule
Area & 
\begin{tabular}[c]{@{}c@{}}Area\\(km\textsuperscript{2})\end{tabular} &
Images & 
\begin{tabular}[c]{@{}c@{}}GSD\\(cm/pixel)\end{tabular} &
Overlap \\
\midrule
NanFang     & 1.24 & 11,961 & 3 & 90\%/80\% \\
YangHaiTang & 1.27 &  9,693 & 3 & 90\%/80\% \\
XiaoXiang   & 1.58 & 20,159 & 3 & 90\%/80\% \\
\bottomrule
\end{tabular}
\endgroup
\end{table}

\begin{figure}[h]
\centering
\includegraphics[width=\linewidth]{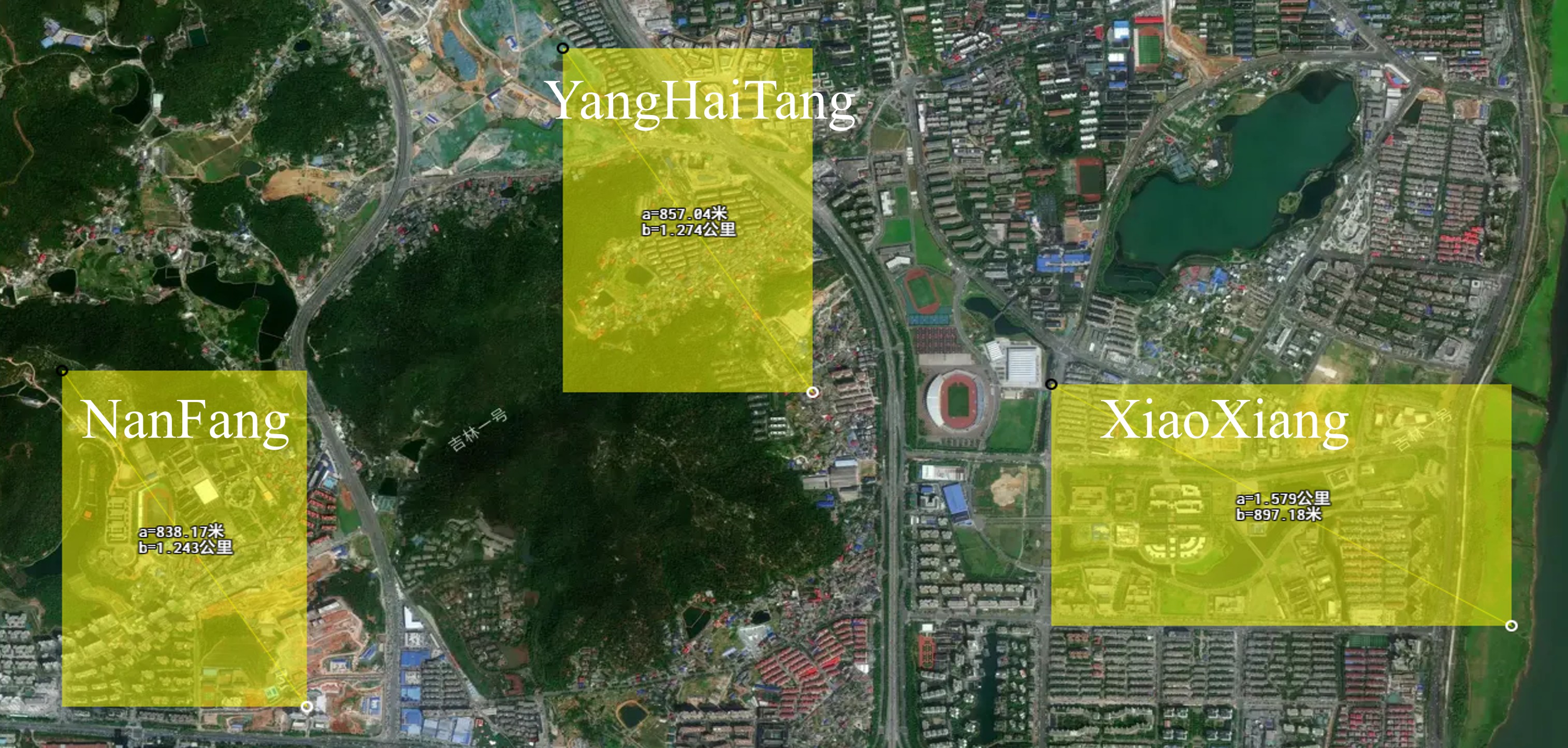}
\caption{Geographic coverage of the three LiDAR-supported UAVFF3D-Real acquisition areas.}
\label{fig:a3d_real_test_scene_coverage}
\end{figure}

\begin{figure}[h]
\centering
\includegraphics[width=\linewidth]{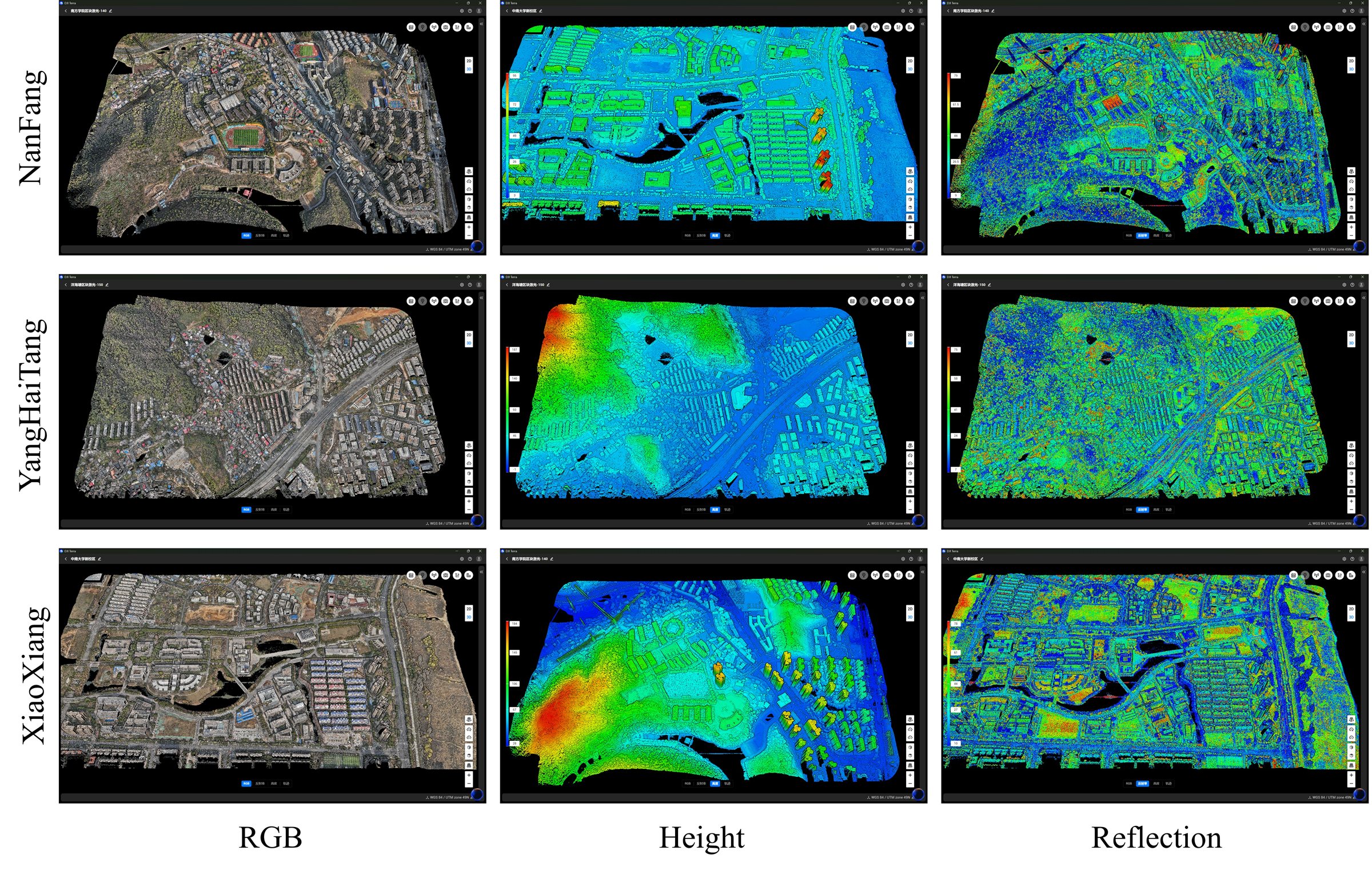}
\caption{LiDAR acquisition results.}
\label{fig:lidar}
\end{figure}

NanFang mainly contains dense low-rise residential buildings and urban blocks.
YangHaiTang contains campus-scale and urban structures with distinct facade and roof variations.
XiaoXiang Campus contains buildings, vegetation, roads, and waterfront areas, making it suitable for evaluating visibility and geometric consistency in complex real scenes.

\subsection{Unified Data Representation and Processing Pipeline}
\label{sec:appendix_unified_pipeline}
UAVFF3D unifies real UAV image blocks, UAV--LiDAR acquisitions, and textured 3D models into the same data format.
Each processed scene contains, when available, RGB images, camera intrinsics, camera poses, camera rays, image-aligned depth maps, valid masks, and scene-level metadata.

For image-only real UAV data, camera poses and dense geometry are obtained through SfM/MVS, after which image-aligned depth maps are rendered.
For LiDAR-supported data, the LiDAR point cloud is first registered to the SfM coordinate system using ICP and is then projected and rendered as image-aligned reference depth.
Synthetic data are rendered from high-quality textured 3D models with explicit control over HFOV, flight altitude, viewing direction, and acquisition pattern.

\subsection{LiDAR--SfM Registration and Reference-Depth Generation}
\label{sec:appendix_lidar_depth}
\label{sec:appendix_lidar_sfm_alignment}

For LiDAR-supported real UAV test scenes, we use the SfM reconstruction as the image-side reference coordinate system and align the independently acquired LiDAR point cloud to this coordinate system.
Specifically, we first perform SfM reconstruction on the UAV images to obtain camera intrinsics, extrinsics, and sparse reconstructed points.
Then, following GauU-Scene, which aligns LiDAR point clouds with SfM/COLMAP reconstructions in a common coordinate system, we transform the LiDAR point cloud into the SfM coordinate system using scene-level initial alignment and further refine the registration using ICP.

After registration, we project the LiDAR point cloud into each image view using the SfM camera parameters and render image-aligned LiDAR-supported reference depth and valid masks through visibility testing.
This reference depth is used for metric evaluation in real UAV test scenes.

In practice, we also reconstruct an SfM/MVS mesh model for each test scene and render mesh depth using the same camera parameters.
Mesh depth is not used as the final evaluation reference, but is used only to filter obvious outliers in the projected LiDAR depth, such as isolated noisy points, occlusion-boundary mismatches, or locally inconsistent points.
The final reference depth is still generated from the registered LiDAR point cloud.

\subsection{Controlled HFOV--Height Settings in UAVFF3D-FA}
\label{sec:appendix_a3d_fa}
UAVFF3D-FA is used to isolate HFOV--height ambiguity.
It contains four synthetic scenes; for each scene, eight nadir acquisition groups are generated with HFOVs of
\(25^\circ\), \(35^\circ\), \(45^\circ\), \(55^\circ\), \(65^\circ\), \(75^\circ\), \(85^\circ\), and \(95^\circ\).
The flight altitude is jointly adjusted within approximately 40--210~m so that different HFOV settings have approximately comparable ground footprints.
UAVFF3D-FA is used only for controlled evaluation and is not involved in any fine-tuning.

\begin{figure}[h]
\centering
\includegraphics[width=\linewidth]{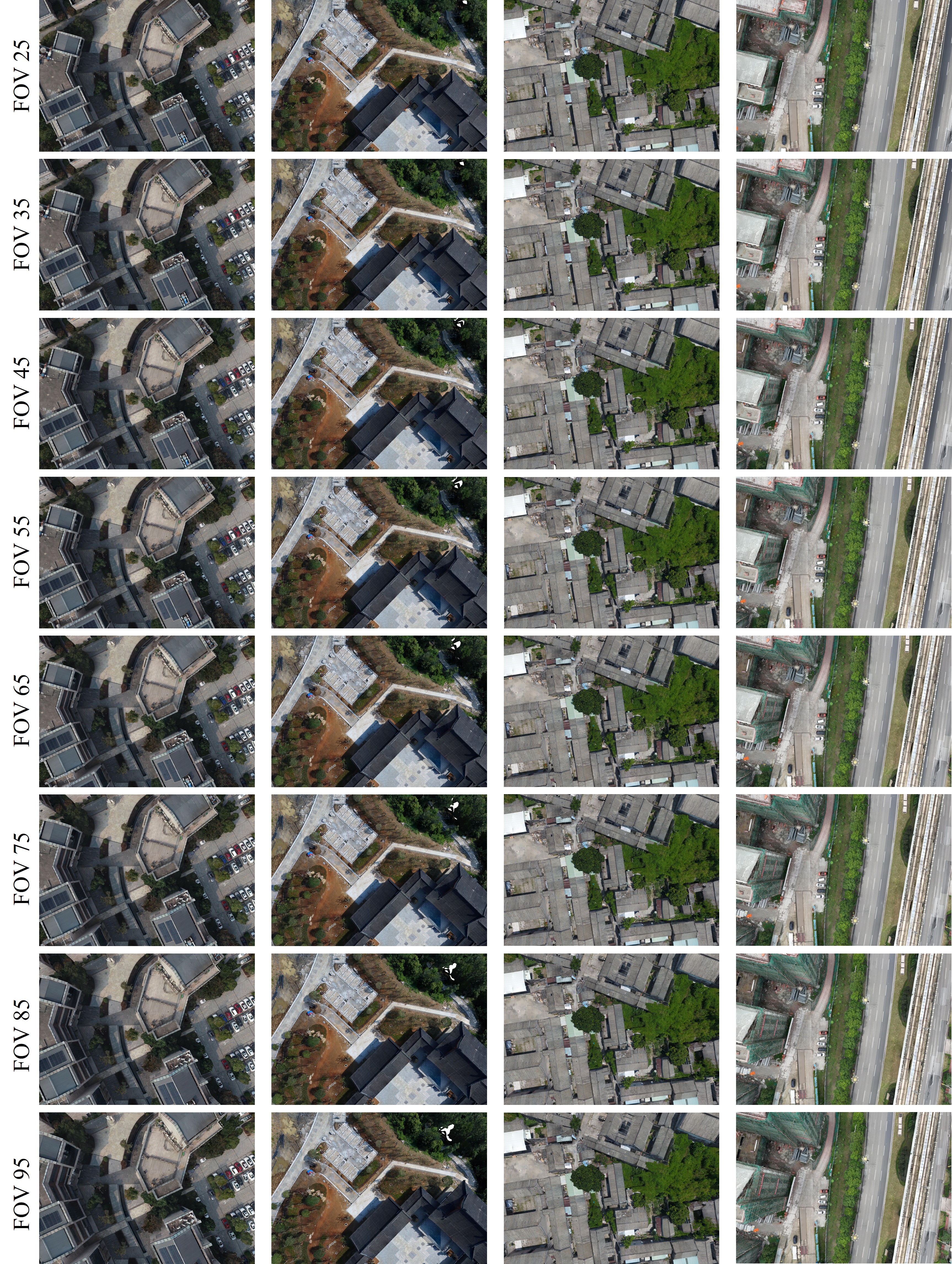}
\caption{Controlled HFOV--height examples in UAVFF3D-FA. Each column corresponds to a scene, and each row shows a different HFOV--height setting. In flat areas dominated by low-rise buildings, images captured under different HFOV values may look almost identical because the ground footprint is approximately preserved. Distinguishable visual differences mainly appear in scenes with high-rise structures, where perspective deformation and vertical geometry reveal HFOV changes.}
\label{fig:a3d_fa_HFOV_height_examples}
\end{figure}

\section{Training, Evaluation, and Implementation Details}
\label{sec:appendix_eval_details}

\subsection{Training and Evaluation Dataset Processing}
\label{sec:appendix_dataset_processing}
The main text provides the overall composition of the training and evaluation data.
Here, we report the camera-geometry ranges of each UAV evaluation set to facilitate reproduction of the data splits and diagnostic settings used in the experiments.
Table~\ref{tab:evaluation_datasets} reports the HFOV and flight-altitude ranges of UseGeo, UrbanScene3D, UAVFF3D-Real, and UAVFF3D-FA.

\begin{table}[h]
\centering
\caption{Detailed camera-geometry coverage of the UAV evaluation datasets. The main text summarizes these HFOV and flight-altitude ranges in a compact form.}
\label{tab:evaluation_datasets}
\begingroup
\TableSetup
\begin{tabular}{@{\hspace{0.4em}}lccc@{\hspace{0.4em}}}
\toprule
Dataset & Acquisition mode & HFOV & Flight altitude (m) \\
\midrule
UrbanScene3D & oblique & \(37^\circ\) & 139, 122 \\
UseGeo & nadir & \(81^\circ\) & 80 \\
UAVFF3D-Real & oblique/nadir & \(36^\circ, 50^\circ, 57^\circ\) & 150, 170, 191 \\
\midrule
UAVFF3D-FA & nadir & \(25^\circ\)--\(95^\circ\) & 40--210 \\
\bottomrule
\end{tabular}
\endgroup
\end{table}

\subsection{Model Inputs and Geometric-Prior Interface}
\label{sec:appendix_prior_interface}
\label{sec:appendix_model_intro}
Table~\ref{tab:baseline_capability} summarizes the input settings supported by each model.
RGB denotes image-only input, C denotes providing camera intrinsics, P denotes providing camera poses, and CP denotes providing both.
VGGT and Pi3 are evaluated only in the RGB-only setting; MapAnything and Pi3X support C/P/CP; DA3 is evaluated with RGB and CP according to its public interface.
We do not report a fine-tuned version of DA3 because its training code and complete fine-tuning pipeline are not publicly available.

\begin{table}[h]
\centering
\caption{Input settings supported by the evaluated feed-forward reconstruction models. RGB denotes image-only inference, C denotes camera intrinsics, P denotes camera poses, and CP denotes jointly using intrinsics and poses.}
\label{tab:baseline_capability}
\begingroup
\TableSetup
\begin{tabular}{@{\hspace{0.4em}}lcccc@{\hspace{0.4em}}}
\toprule
Model & RGB & C & P & CP \\
\midrule
VGGT & \checkmark &  &  &   \\
Pi3 & \checkmark &  &  &   \\
MapAnything & \checkmark & \checkmark & \checkmark & \checkmark \\
Pi3X & \checkmark & \checkmark & \checkmark & \checkmark \\
DA3 & \checkmark &  &  & \checkmark \\
\bottomrule
\end{tabular}
\endgroup
\end{table}

\section{Supplementary Experimental Results}
\label{sec:appendix_additional_results}

\subsection{Supplementary Table Results}
\label{sec:appendix_table_results}
\label{sec:appendix_full_results}
\label{sec:appendix_fa_results}
\label{sec:appendix_prior_view_results}

This subsection summarizes the supplementary quantitative results, including the complete average and dataset-level results, the controlled UAVFF3D-FA diagnostic results, and the prior-aware oblique/nadir breakdown.

\begin{table}[t]
\centering
\caption{Complete average results over the four UAV test datasets. Scores are averaged over 8-, 16-, 24-, and 32-view inputs; all metrics are lower-is-better.}
\label{tab:appendix_results_averaged}
\begingroup
\TableSetup
\begin{tabular}{@{\hspace{0.4em}}lcccccc@{\hspace{0.4em}}}
\toprule
Model & Input & AbsRel$\downarrow$ & CD$\downarrow$ & Ray$\downarrow$ & ATE$\downarrow$ & Rot.$\downarrow$ \\
\midrule
\multicolumn{7}{l}{\textit{Zero-shot models}} \\
\midrule
MapAnything & RGB & 0.044 & 6.00 & 2.90 & 57.50 & 13.33 \\
MapAnything & C & 0.043 & 0.96 & 2.51 & 38.52 & 12.32 \\
MapAnything & P & 0.105 & 5.79 & 5.66 & 63.97 & 7.34 \\
MapAnything & CP & 0.099 & 0.66 & 5.15 & 44.72 & 7.73 \\
Pi3X & RGB & 0.035 & 7.73 & 2.24 & 56.19 & 11.72 \\
Pi3X & C & 0.032 & 0.24 & 2.08 & 23.17 & 9.27 \\
Pi3X & P & 0.032 & 7.44 & 2.14 & 48.23 & 6.46 \\
Pi3X & CP & 0.029 & \textbf{0.23} & 2.04 & 13.48 & 4.38 \\
DA3 & RGB & 0.037 & 6.58 & 2.94 & 56.21 & 14.79 \\
DA3 & CP & 0.039 & 2.51 & 3.11 & 27.88 & 11.86 \\
\midrule
\multicolumn{7}{l}{\textit{Prior-aware evaluation after UAV-domain adaptation}} \\
\midrule
Mapa-FT & RGB & 0.037 & 2.51 & 1.91 & 29.08 & 6.10 \\
Mapa-FT & C & 0.033 & 0.38 & 1.86 & 15.86 & 5.55 \\
Mapa-FT & P & 0.025 & 2.38 & 1.35 & 17.61 & 1.61 \\
Mapa-FT & CP & 0.021 & 0.30 & \textbf{1.14} & \textbf{3.43} & \textbf{1.06} \\
Pi3X-FT & RGB & 0.026 & 1.87 & 1.44 & 18.70 & 3.40 \\
Pi3X-FT & C & 0.020 & 0.25 & 1.63 & 8.13 & 3.14 \\
Pi3X-FT & P & 0.021 & 1.93 & 1.77 & 16.50 & 1.98 \\
Pi3X-FT & CP & \textbf{0.018} & 0.26 & 1.89 & 4.80 & 1.62 \\
\bottomrule
\end{tabular}
\endgroup
\end{table}

\begin{table}[t]
\centering
\caption{Complete UrbanScene3D results for oblique urban reconstruction. All metrics are lower-is-better.}
\label{tab:appendix_results_urbanscene3d}
\begingroup
\TableSetup
\begin{tabular}{@{\hspace{0.4em}}lcccccc@{\hspace{0.4em}}}
\toprule
Model & Input & AbsRel$\downarrow$ & CD$\downarrow$ & Ray$\downarrow$ & ATE$\downarrow$ & Rot.$\downarrow$ \\
\midrule
\multicolumn{7}{l}{\textit{Zero-shot models}} \\
\midrule
VGGT & RGB & 0.069 & 4.75 & 8.47 & 77.78 & 26.00 \\
Pi3 & RGB & 0.055 & 4.85 & 12.91 & 85.50 & 25.86 \\
Mapa & RGB & 0.059 & 4.96 & 8.25 & 81.12 & 25.57 \\
Mapa & C & 0.063 & 4.70 & 0.99 & 58.22 & 24.07 \\
Mapa & P & 0.072 & 5.97 & 7.86 & 86.95 & 6.74 \\
Mapa & CP & 0.086 & 5.73 & 0.62 & 63.07 & 7.46 \\
Pi3X & RGB & 0.046 & 4.04 & 12.92 & 78.40 & 26.11 \\
Pi3X & C & 0.055 & 3.85 & 0.18 & 44.49 & 23.83 \\
Pi3X & P & 0.048 & 3.58 & 12.23 & 68.98 & 12.98 \\
Pi3X & CP & 0.051 & 3.72 & 0.15 & 25.74 & 9.98 \\
DA3 & RGB & 0.057 & 5.74 & 9.37 & 88.11 & 35.04 \\
DA3 & CP & 0.054 & 5.01 & 3.15 & 37.93 & 24.98 \\
\midrule
\multicolumn{7}{l}{\textit{UAV-domain-adapted models}} \\
\midrule
VGGT-FT & RGB & 0.042 & 2.99 & 1.85 & 28.84 & 9.35 \\
Pi3-FT & RGB & 0.041 & 3.20 & 1.03 & 21.63 & 7.47 \\
Mapa-FT & RGB & 0.049 & 3.40 & 2.43 & 39.31 & 12.90 \\
Mapa-FT & C & 0.048 & 3.40 & 0.24 & 27.86 & 11.31 \\
Mapa-FT & P & 0.036 & 1.95 & 1.67 & 14.55 & 2.26 \\
Mapa-FT & CP & 0.029 & \textbf{1.59} & 0.14 & \textbf{4.18} & \textbf{1.36} \\
Pi3X-FT & RGB & 0.038 & 2.78 & 1.24 & 24.03 & 8.39 \\
Pi3X-FT & C & 0.035 & 3.21 & \textbf{0.13} & 18.53 & 8.11 \\
Pi3X-FT & P & 0.033 & 2.52 & 1.21 & 17.36 & 3.82 \\
Pi3X-FT & CP & \textbf{0.028} & 2.77 & 0.19 & 9.39 & 3.43 \\
\bottomrule
\end{tabular}
\endgroup
\end{table}

\begin{table}[t]
\centering
\caption{Complete UseGeo results for wide-HFOV nadir UAV evaluation. All metrics are lower-is-better.}
\label{tab:appendix_results_usegeo}
\begingroup
\TableSetup
\begin{tabular}{@{\hspace{0.4em}}lcccccc@{\hspace{0.4em}}}
\toprule
Model & Input & AbsRel$\downarrow$ & CD$\downarrow$ & Ray$\downarrow$ & ATE$\downarrow$ & Rot.$\downarrow$ \\
\midrule
\multicolumn{7}{l}{\textit{Zero-shot models}} \\
\midrule
VGGT & RGB & 0.027 & 1.61 & 1.99 & 8.32 & 2.10 \\
Pi3 & RGB & 0.025 & 1.79 & 4.51 & 15.98 & 2.42 \\
Mapa & RGB & 0.029 & 1.57 & 2.12 & 8.80 & 2.73 \\
Mapa & C & 0.028 & 1.17 & 0.83 & 4.84 & 2.55 \\
Mapa & P & 0.143 & 6.17 & 2.04 & 22.01 & 6.78 \\
Mapa & CP & 0.128 & 5.26 & 0.70 & 20.96 & 6.30 \\
Pi3X & RGB & 0.028 & 1.41 & 1.94 & 8.10 & 2.16 \\
Pi3X & C & 0.020 & 1.20 & 0.44 & 3.22 & 1.90 \\
Pi3X & P & 0.023 & \textbf{0.93} & 1.85 & 7.13 & 1.28 \\
Pi3X & CP & 0.017 & 1.06 & \textbf{0.43} & 2.17 & 1.09 \\
DA3 & RGB & 0.030 & 1.72 & 3.23 & 10.32 & 4.15 \\
DA3 & CP & 0.033 & 1.78 & 2.00 & 7.56 & 4.83 \\
\midrule
\multicolumn{7}{l}{\textit{UAV-domain-adapted models}} \\
\midrule
VGGT-FT & RGB & 0.018 & 0.99 & 0.62 & 2.92 & 1.43 \\
Pi3-FT & RGB & 0.019 & 1.27 & 0.65 & 3.20 & 1.30 \\
Mapa-FT & RGB & 0.025 & 1.15 & 0.76 & 4.54 & 2.33 \\
Mapa-FT & C & 0.025 & 1.14 & 0.57 & 3.85 & 2.28 \\
Mapa-FT & P & 0.019 & 0.99 & 0.75 & 2.28 & \textbf{0.72} \\
Mapa-FT & CP & 0.019 & 0.99 & 0.51 & \textbf{1.61} & 0.81 \\
Pi3X-FT & RGB & 0.020 & 0.93 & 0.90 & 3.99 & 1.28 \\
Pi3X-FT & C & 0.017 & 1.11 & 0.46 & 2.47 & 1.33 \\
Pi3X-FT & P & 0.017 & 1.35 & 0.84 & 3.11 & 0.85 \\
Pi3X-FT & CP & \textbf{0.015} & 1.49 & 0.46 & 1.71 & 0.80 \\
\bottomrule
\end{tabular}
\endgroup
\end{table}

\begin{table}[t]
\centering
\caption{Complete UAVFF3D-Real results on LiDAR-supported real UAV evaluation scenes. All metrics are lower-is-better.}
\label{tab:appendix_results_a3d_real}
\begingroup
\TableSetup
\begin{tabular}{@{\hspace{0.4em}}lcccccc@{\hspace{0.4em}}}
\toprule
Model & Input & AbsRel$\downarrow$ & CD$\downarrow$ & Ray$\downarrow$ & ATE$\downarrow$ & Rot.$\downarrow$ \\
\midrule
\multicolumn{7}{l}{\textit{Zero-shot models}} \\
\midrule
VGGT & RGB & 0.058 & 3.38 & 7.67 & 89.45 & 16.98 \\
Pi3 & RGB & 0.050 & 3.47 & 10.51 & 98.72 & 16.16 \\
Mapa & RGB & 0.064 & 3.88 & 6.55 & 97.46 & 22.74 \\
Mapa & C & 0.060 & 3.48 & 0.95 & 78.25 & 20.06 \\
Mapa & P & 0.107 & 6.98 & 6.22 & 91.59 & 8.65 \\
Mapa & CP & 0.093 & 6.57 & 0.56 & 63.88 & 9.91 \\
Pi3X & RGB & 0.055 & 2.70 & 9.58 & 98.26 & 17.52 \\
Pi3X & C & 0.046 & 2.71 & 0.23 & 42.77 & 10.55 \\
Pi3X & P & 0.047 & 3.39 & 9.24 & 76.76 & 10.47 \\
Pi3X & CP & 0.040 & 2.83 & \textbf{0.23} & 24.06 & 5.82 \\
DA3 & RGB & 0.050 & 3.59 & 7.58 & 87.38 & 18.45 \\
DA3 & CP & 0.058 & 5.04 & 2.80 & 50.99 & 16.13 \\
\midrule
\multicolumn{7}{l}{\textit{UAV-domain-adapted models}} \\
\midrule
VGGT-FT & RGB & 0.038 & 1.77 & 2.28 & 29.32 & 3.86 \\
Pi3-FT & RGB & 0.031 & 1.93 & 1.05 & 16.61 & 2.47 \\
Mapa-FT & RGB & 0.049 & 2.32 & 1.92 & 42.51 & 7.18 \\
Mapa-FT & C & 0.039 & 2.23 & 0.27 & 26.17 & 6.63 \\
Mapa-FT & P & 0.027 & 1.73 & 1.58 & 19.15 & 1.99 \\
Mapa-FT & CP & 0.021 & \textbf{1.40} & 0.23 & \textbf{4.87} & \textbf{1.23} \\
Pi3X-FT & RGB & 0.033 & 1.48 & 1.41 & 20.01 & 2.92 \\
Pi3X-FT & C & 0.020 & 1.63 & 0.27 & 9.40 & 2.35 \\
Pi3X-FT & P & 0.024 & 2.32 & 1.45 & 16.76 & 2.31 \\
Pi3X-FT & CP & \textbf{0.018} & 2.33 & 0.28 & 6.40 & 1.65 \\
\bottomrule
\end{tabular}
\endgroup
\end{table}

\begin{table}[t]
\centering
\caption{Complete prior-aware oblique/nadir results averaged over the evaluated UAV datasets. Each model/input setting reports oblique and nadir rows separately. All metrics are lower-is-better.}
\label{tab:appendix_oblique_nadir_prior_results}
\begingroup
\TableSetup
\begin{tabular}{@{\hspace{0.4em}}lccccccc@{\hspace{0.4em}}}
\toprule
Model & Input & View & AbsRel$\downarrow$ & CD$\downarrow$ & Ray$\downarrow$ & ATE$\downarrow$ & Rot.$\downarrow$ \\
\midrule
VGGT & RGB & Oblique & 0.094 & 5.42 & 8.35 & 109.28 & 31.89 \\
VGGT & RGB & Nadir & 0.023 & 2.02 & 6.12 & 38.10 & 6.32 \\
Pi3 & RGB & Oblique & 0.078 & 5.80 & 12.50 & 121.52 & 33.00 \\
Pi3 & RGB & Nadir & 0.023 & 2.29 & 8.09 & 44.71 & 6.86 \\
Mapa & RGB & Oblique & 0.088 & 6.51 & 8.22 & 128.33 & 39.02 \\
Mapa & RGB & Nadir & 0.029 & 1.79 & 5.41 & 35.36 & 5.12 \\
Mapa & C & Oblique & 0.088 & 6.26 & 1.10 & 109.96 & 34.23 \\
Mapa & C & Nadir & 0.028 & 1.26 & 0.87 & 13.40 & 5.04 \\
Mapa & P & Oblique & 0.177 & 9.81 & 7.72 & 114.09 & 17.64 \\
Mapa & P & Nadir & 0.100 & 5.39 & 5.32 & 59.60 & 6.72 \\
Mapa & CP & Oblique & 0.153 & 10.13 & 0.61 & 72.72 & 16.19 \\
Mapa & CP & Nadir & 0.095 & 4.89 & 0.65 & 42.93 & 6.91 \\
Pi3X & RGB & Oblique & 0.078 & 5.60 & 12.49 & 124.54 & 35.89 \\
Pi3X & RGB & Nadir & 0.023 & 1.89 & 6.77 & 40.56 & 10.11 \\
Pi3X & C & Oblique & 0.070 & 4.62 & 0.22 & 69.22 & 24.39 \\
Pi3X & C & Nadir & 0.020 & 1.44 & 0.23 & 8.01 & 7.07 \\
Pi3X & P & Oblique & 0.077 & 6.45 & 11.94 & 94.63 & 22.98 \\
Pi3X & P & Nadir & 0.021 & 1.11 & 6.51 & 37.88 & 2.57 \\
Pi3X & CP & Oblique & 0.064 & 5.08 & \textbf{0.20} & 39.50 & 12.63 \\
Pi3X & CP & Nadir & 0.017 & 0.99 & \textbf{0.23} & 4.08 & 1.65 \\
DA3 & RGB & Oblique & 0.069 & 5.84 & 9.44 & 114.68 & 32.95 \\
DA3 & RGB & Nadir & 0.025 & 2.05 & 5.86 & 36.80 & 10.81 \\
DA3 & CP & Oblique & 0.079 & 7.88 & 3.42 & 72.22 & 27.32 \\
DA3 & CP & Nadir & 0.030 & 2.14 & 2.13 & 17.31 & 9.12 \\
\midrule
VGGT-FT & RGB & Oblique & 0.051 & 2.86 & 1.52 & 30.98 & 7.79 \\
VGGT-FT & RGB & Nadir & 0.019 & 1.19 & 2.40 & 18.86 & 3.07 \\
Pi3-FT & RGB & Oblique & 0.042 & 2.89 & 0.78 & 20.12 & 5.31 \\
Pi3-FT & RGB & Nadir & 0.021 & 1.49 & 1.58 & 13.69 & 2.88 \\
Mapa-FT & RGB & Oblique & 0.061 & 3.67 & 1.70 & 49.61 & 13.06 \\
Mapa-FT & RGB & Nadir & 0.028 & 1.23 & 2.93 & 22.27 & 3.64 \\
Mapa-FT & C & Oblique & 0.054 & 3.66 & 0.29 & 37.47 & 11.66 \\
Mapa-FT & C & Nadir & 0.025 & 1.15 & 0.35 & 7.48 & 3.36 \\
Mapa-FT & P & Oblique & 0.036 & 2.20 & 1.00 & 13.13 & 2.21 \\
Mapa-FT & P & Nadir & 0.021 & 1.03 & 2.94 & 21.09 & 1.57 \\
Mapa-FT & CP & Oblique & 0.028 & \textbf{1.79} & 0.20 & \textbf{5.10} & \textbf{1.29} \\
Mapa-FT & CP & Nadir & 0.018 & \textbf{0.88} & 0.29 & 2.98 & \textbf{1.07} \\
Pi3X-FT & RGB & Oblique & 0.044 & 2.53 & 0.94 & 23.50 & 6.29 \\
Pi3X-FT & RGB & Nadir & 0.019 & 1.10 & 2.21 & 17.54 & 2.95 \\
Pi3X-FT & C & Oblique & 0.032 & 2.86 & 0.22 & 17.87 & 5.94 \\
Pi3X-FT & C & Nadir & 0.017 & 1.24 & 0.23 & 4.77 & 2.58 \\
Pi3X-FT & P & Oblique & 0.034 & 2.84 & 0.74 & 15.18 & 3.70 \\
Pi3X-FT & P & Nadir & 0.016 & 1.40 & 2.43 & 18.07 & 1.42 \\
Pi3X-FT & CP & Oblique & \textbf{0.027} & 3.00 & 0.27 & 10.28 & 3.11 \\
Pi3X-FT & CP & Nadir & \textbf{0.014} & 1.48 & 0.23 & \textbf{2.73} & 1.12 \\
\bottomrule
\end{tabular}
\endgroup
\end{table}

\begin{table}[t]
\centering
\caption{Complete prior-aware UAVFF3D-FA results averaged over controlled HFOV settings. RGB/C/P/CP denote image-only, intrinsics, poses, and both priors, respectively; all metrics are lower-is-better.}
\label{tab:appendix_a3dfa_prior_results}
\begingroup
\TableSetup
\begin{tabular}{@{\hspace{0.4em}}lcccccc@{\hspace{0.4em}}}
\toprule
Model & Input & AbsRel$\downarrow$ & CD$\downarrow$ & Ray$\downarrow$ & ATE$\downarrow$ & Rot.$\downarrow$ \\
\midrule
\multicolumn{7}{l}{\textit{Zero-shot models}} \\
\midrule
VGGT & RGB & 0.013 & 1.04 & 6.80 & 41.62 & 1.33 \\
Pi3 & RGB & 0.012 & 1.31 & 6.73 & 41.36 & 1.21 \\
Mapa & RGB & 0.021 & 1.19 & 7.06 & 42.64 & 2.28 \\
Mapa & C & 0.021 & 0.70 & 1.06 & 12.79 & 2.62 \\
Mapa & P & 0.096 & 3.52 & 7.05 & 55.35 & 7.18 \\
Mapa & CP & 0.089 & 3.05 & 0.75 & 30.98 & 7.25 \\
Pi3X & RGB & 0.011 & 0.81 & 6.47 & 40.00 & 1.09 \\
Pi3X & C & 0.009 & 0.55 & \textbf{0.12} & 2.19 & 0.78 \\
Pi3X & P & 0.010 & 0.65 & 6.46 & 40.03 & 1.09 \\
Pi3X & CP & \textbf{0.009} & \textbf{0.55} & 0.14 & 1.96 & 0.65 \\
DA3 & RGB & 0.011 & 0.73 & 6.15 & 39.02 & 1.50 \\
DA3 & CP & 0.012 & 0.61 & 2.09 & 15.04 & 1.49 \\
\midrule
\multicolumn{7}{l}{\textit{UAV-domain-adapted models}} \\
\midrule
VGGT-FT & RGB & 0.013 & 0.59 & 3.13 & 21.20 & 1.23 \\
Pi3-FT & RGB & 0.011 & 0.68 & 2.74 & 16.65 & 1.06 \\
Mapa-FT & RGB & 0.025 & 0.80 & 4.93 & 29.97 & 2.00 \\
Mapa-FT & C & 0.020 & 0.67 & 0.43 & 5.57 & 1.97 \\
Mapa-FT & P & 0.018 & 0.72 & 5.51 & 34.46 & 1.45 \\
Mapa-FT & CP & 0.013 & 0.56 & 0.34 & 3.05 & 0.83 \\
Pi3X-FT & RGB & 0.011 & 0.58 & 3.93 & 26.79 & 1.01 \\
Pi3X-FT & C & 0.009 & 0.59 & 0.14 & 2.11 & 0.77 \\
Pi3X-FT & P & 0.010 & 0.91 & 4.21 & 28.78 & 0.94 \\
Pi3X-FT & CP & 0.009 & 0.96 & 0.14 & \textbf{1.72} & \textbf{0.61} \\
\bottomrule
\end{tabular}
\endgroup
\end{table}


\subsection{Supplementary Visualization Results}
\label{sec:appendix_visualization_results}
\label{sec:appendix_dataset_visualization}

\begin{figure*}[h]
\centering
\includegraphics[width=0.95\linewidth]{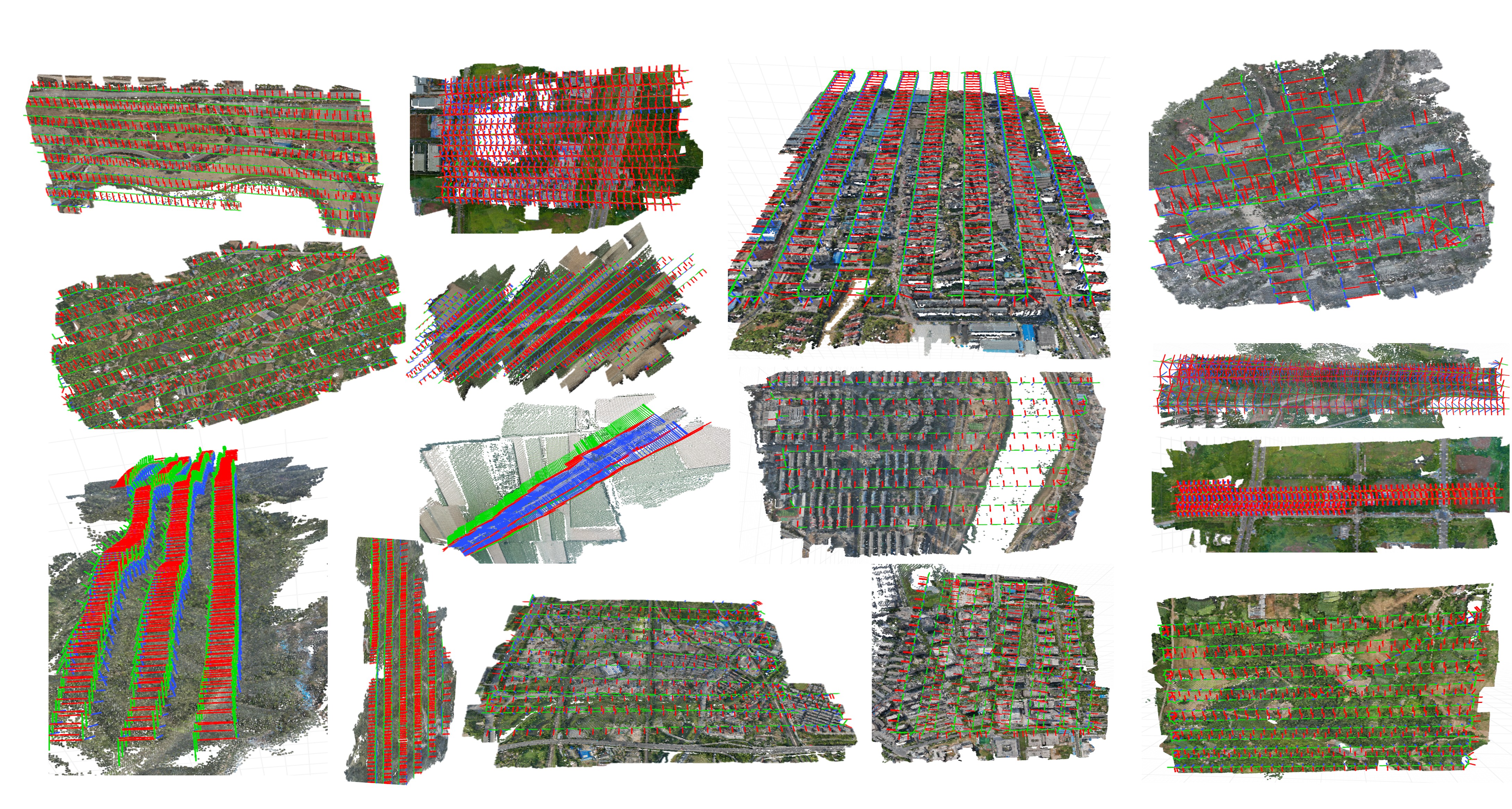}
\caption{Representative scene visualization of UAVFF3D-Real. 
These examples illustrate the diversity of real UAV acquisition in UAVFF3D.
}
\label{fig:a3d_real_vis}
\end{figure*}

\begin{figure*}[h]
\centering
\includegraphics[width=0.95\linewidth]{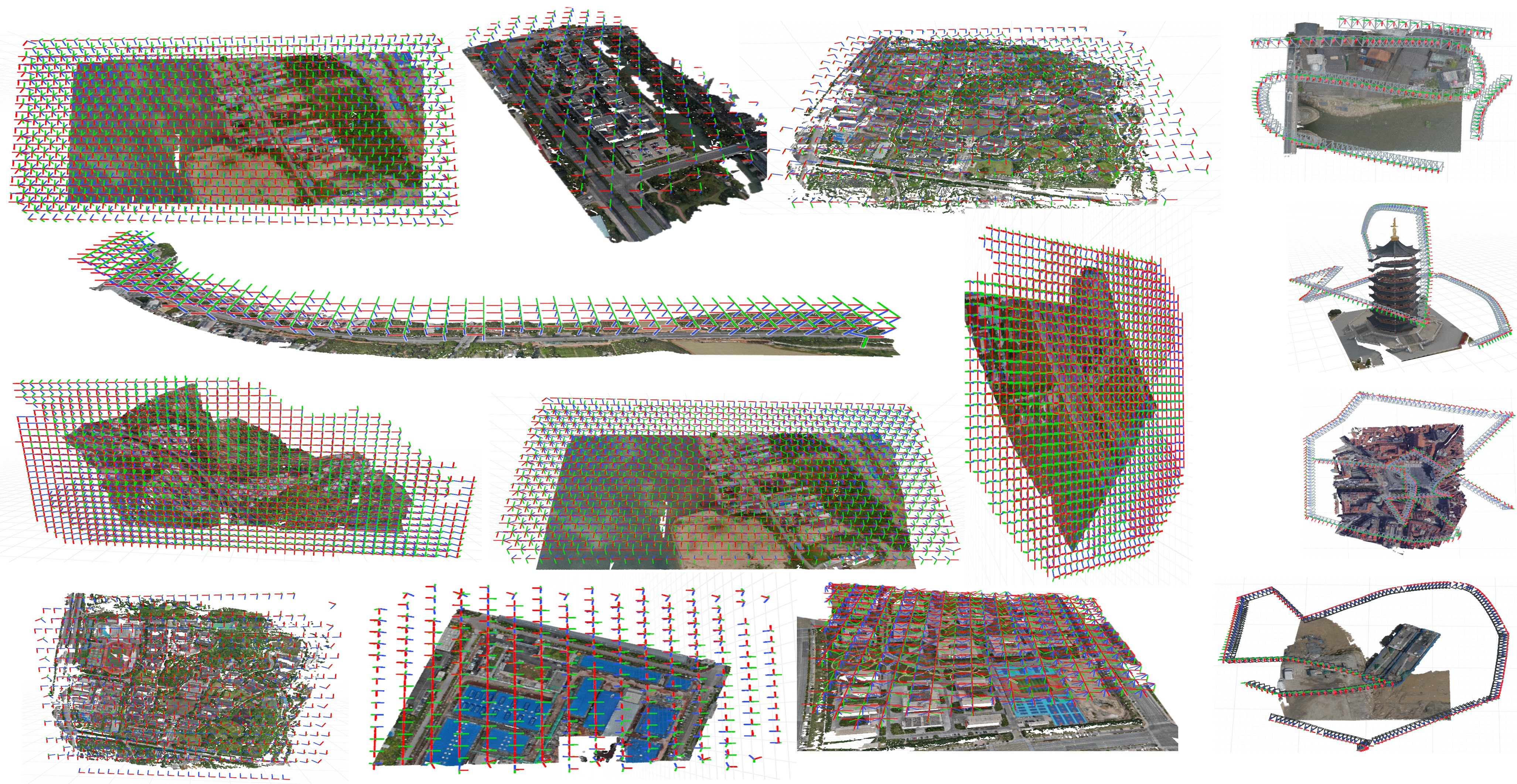}
\caption{Representative scene visualization of UAVFF3D-Syn. 
The synthetic scenes cover diverse UAV camera trajectories and scene layouts. 
}
\label{fig:a3d_syn_vis}
\end{figure*}

\begin{figure*}[t]
\centering
\includegraphics[width=0.95\textwidth]{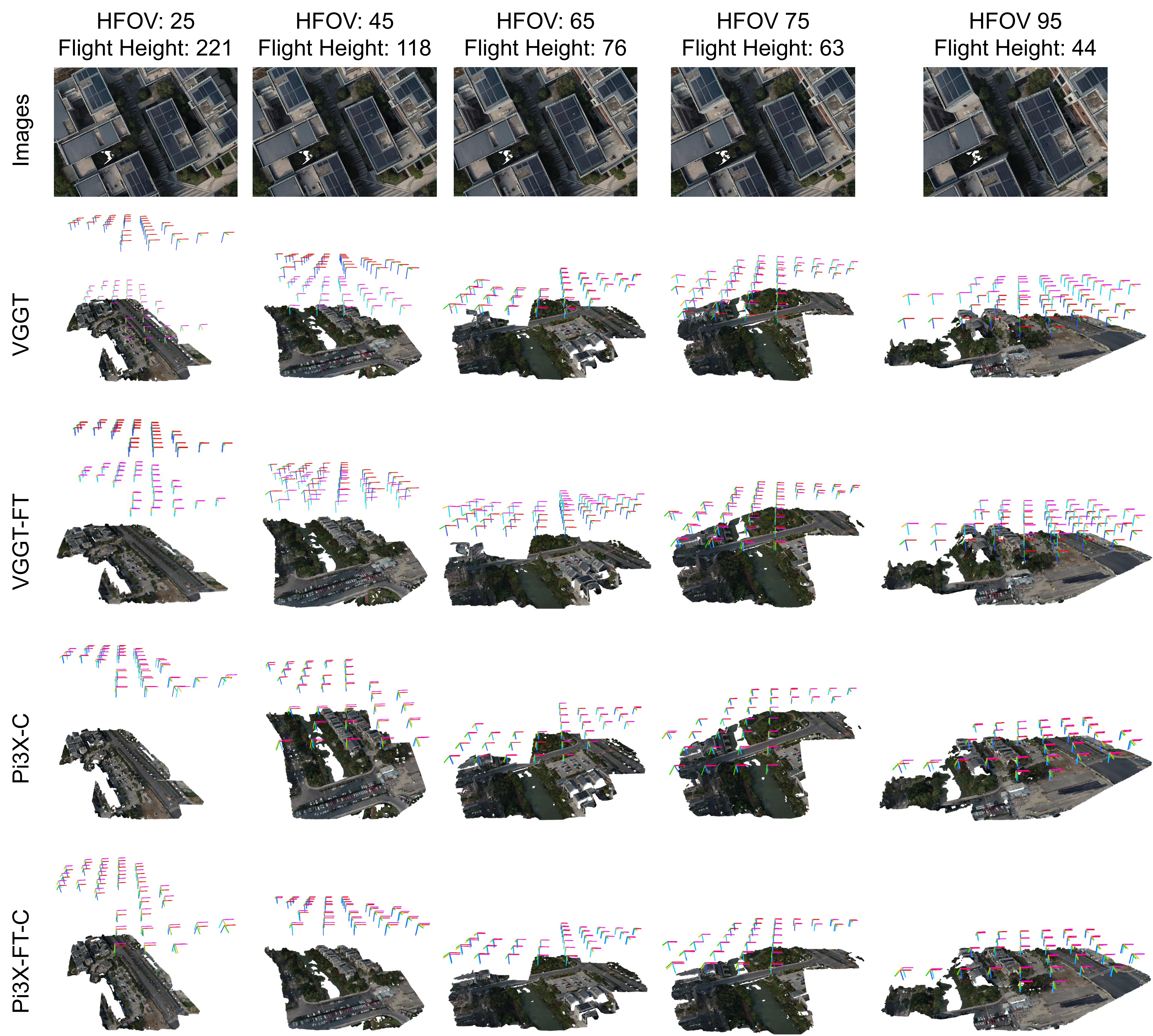}
\caption{Qualitative visualization results on UAVFF3D-FA. The examples show reconstruction behavior under controlled HFOV--height settings.}
\label{fig:uavff3d_fa_vis}
\end{figure*}

\begin{figure*}[t]
\centering
\includegraphics[width=0.95\textwidth]{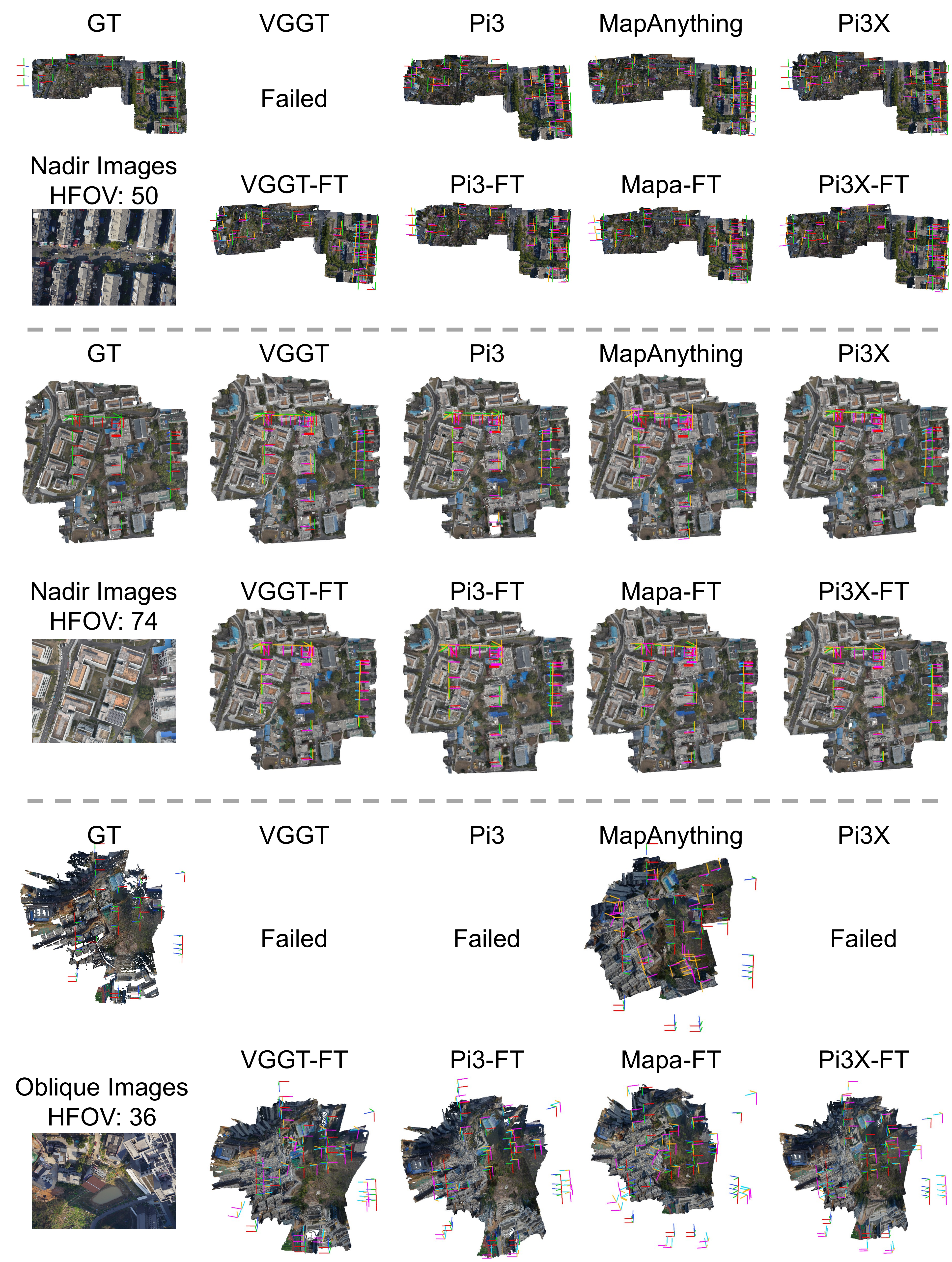}
\caption{Qualitative visualization results on UAVFF3D-Real. The examples show reconstruction outputs on real UAV scenes from the UAVFF3D-Real test split.}
\label{fig:uavff3d_real_vis}
\end{figure*}

\begin{figure*}[t]
\centering
\includegraphics[width=0.95\textwidth]{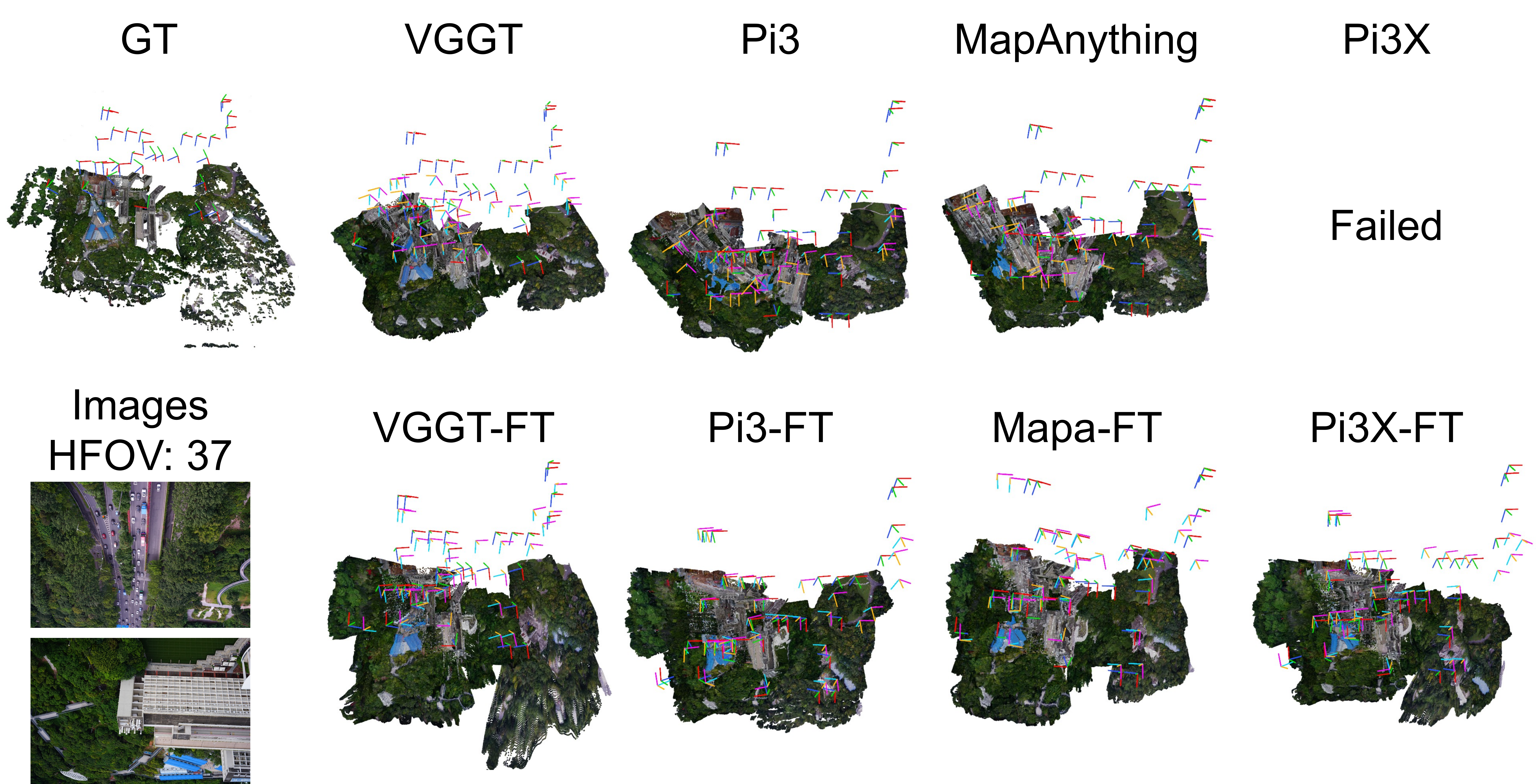}
\caption{Qualitative visualization results on UrbanScene3D. The examples illustrate feed-forward reconstruction under oblique urban UAV acquisition.}
\label{fig:urbanscene3d_vis}
\end{figure*}

\begin{figure*}[t]
\centering
\includegraphics[width=0.95\textwidth]{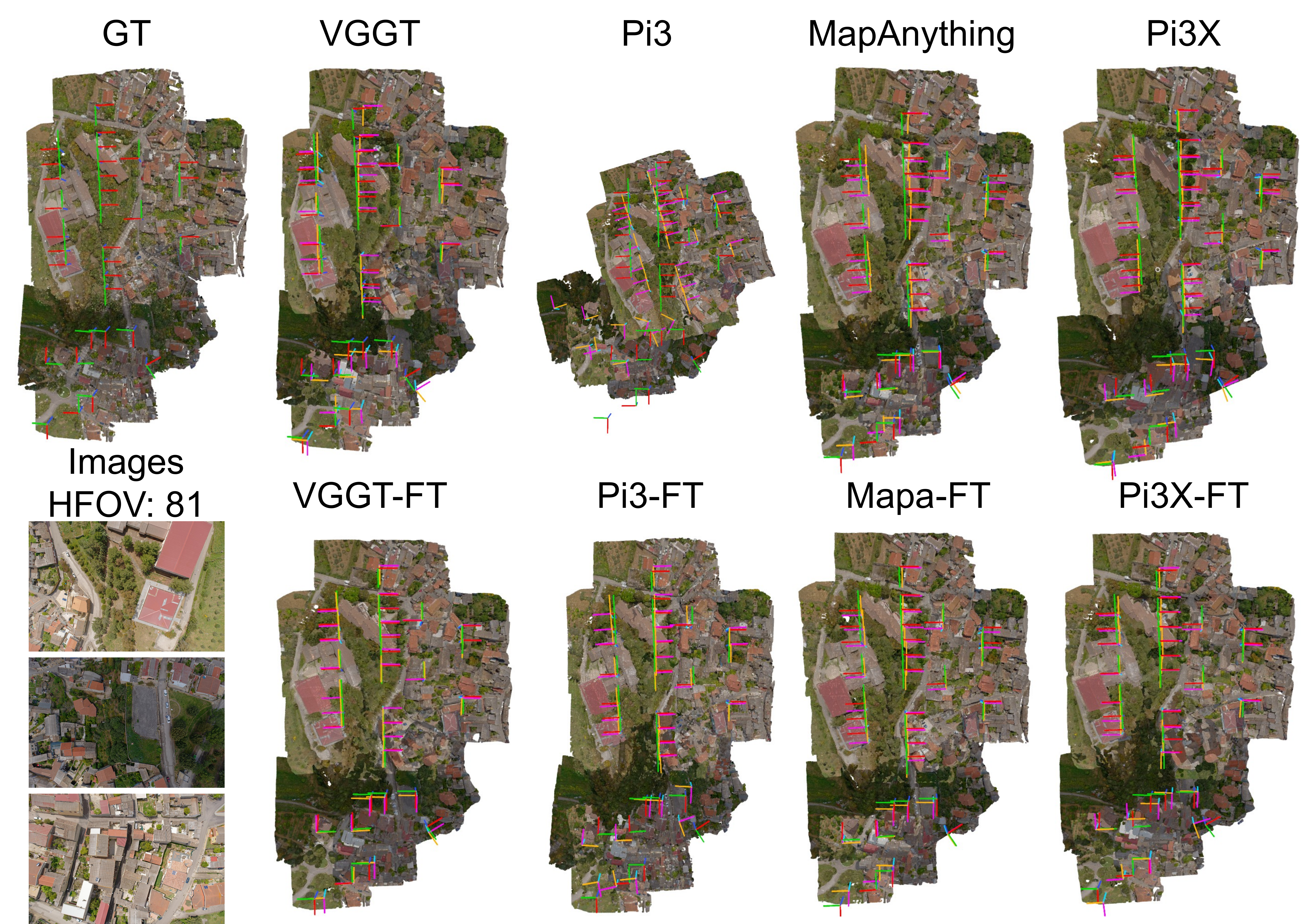}
\caption{Qualitative visualization results on UseGeo. The examples illustrate reconstruction performance on nadir-view UAV scenes.}
\label{fig:usegeo_vis}
\end{figure*}

This subsection provides representative dataset visualizations and qualitative reconstruction results.
Figures~\ref{fig:a3d_real_vis} and~\ref{fig:a3d_syn_vis} show representative examples from UAVFF3D-Real and UAVFF3D-Syn, respectively.
Figure~\ref{fig:a3d_fa_HFOV_height_examples} in Appendix~\ref{sec:appendix_a3d_fa} shows the controlled HFOV--height settings of UAVFF3D-FA.
Figures~\ref{fig:uavff3d_fa_vis}, \ref{fig:uavff3d_real_vis}, \ref{fig:urbanscene3d_vis}, and \ref{fig:usegeo_vis} provide qualitative visualization results for the four UAV test datasets.

\end{document}